\documentclass[lettersize,journal]{IEEEtran}
\usepackage{algorithmic}
\usepackage{algorithm}
\usepackage{array}
\usepackage[caption=false,font=normalsize,labelfont=sf,textfont=sf]{subfig}
\usepackage{textcomp}
\usepackage{stfloats}
\usepackage{url}
\usepackage{verbatim}
\usepackage{graphicx}
\usepackage{cite}

\graphicspath{ {./figures/}} 
\usepackage[T1]{fontenc}

\usepackage{listings}
\usepackage{amsmath,amssymb,amsfonts,amsthm}
\newtheorem{theorem}{Theorem}
\usepackage{mathtools} 

\graphicspath{ {./figures/} }
\usepackage{textcomp}

\usepackage{url}

\usepackage[dvipsnames]{xcolor}

\begin{document}

\title{{\em gc}DLSeg: Integrating Graph-cut into Deep Learning for Binary Semantic Segmentation}

\author{Hui Xie \and Weiyu Xu \and Ya Xing Wang (MD) \and John Buatti (MD) \and  Xiaodong Wu}

\markboth{IEEE TRANSACTIONS ON PATTERN ANALYSIS AND MACHINE INTELLIGENCE, VOL. **, NO. **, **DECEMBER** 2023}%
{Shell \MakeLowercase{\textit{et al.}}: Integrate Min-cut Algorithms in Deep Learning for Binary Semantic Segmentation}


\maketitle

\begin{abstract}
Binary semantic segmentation in computer vision is a fundamental problem. As a model-based segmentation method, the graph-cut approach was one of the most successful binary segmentation methods thanks to its global optimality guarantee of the solutions and its practical polynomial-time complexity. Recently, many deep learning (DL) based methods have been developed for this task and yielded remarkable performance, resulting in a paradigm shift in this field. To combine the strengths of both approaches, we propose in this study to integrate the graph-cut approach into a deep learning network for end-to-end learning. Unfortunately, backward propagation through the graph-cut module in the DL network is challenging due to the combinatorial nature of the graph-cut algorithm. To tackle this challenge, we propose a novel residual graph-cut loss and a quasi-residual connection, enabling the backward propagation of the gradients of the residual graph-cut loss for effective feature learning guided by the graph-cut segmentation model.  In the inference phase, globally optimal segmentation is achieved with respect to the graph-cut energy defined on the optimized image features learned from DL networks. Experiments on the public AZH chronic wound data set and the pancreas cancer data set from the medical segmentation decathlon (MSD) demonstrated promising segmentation accuracy, and improved robustness against adversarial attacks. 
	
\end{abstract}


\section{Introduction} 

\IEEEPARstart{I}{mage} semantic segmentation, which partitions images into multiple segments on the pixel level, plays a fundamental role in computer vision applications~\cite{minaee_dlsegmentationsurvery_2021}, such as scene understanding, remote sensing, autopilot, medical image analysis, robotic perception, video surveillance, augmented reality, and image compression.

Prior to the revolutionary rise of deep learning, the graph-cut method~\cite{boykov_graphcut_2001,boykov_graphcut_ieee_2001,boykov_maxflow_2004} had been one of the major image segmentation approaches~\cite{vese2002multiphase,li2010distance,li2011level}. It was proposed by Greig et al. ~\cite{greig_mpe_1989} and Boykov et al.~\cite{boykov_maxflow_2004} to formulate binary semantic segmentation as a minimum $s$-$t$ cut problem in an associated graph. The graph-cut method is ubiquitous in computer vision as a large variety of computer vision problems can be formulated as a min-cut/max-flow problem~\cite{jensen_mincutreview_2022}. It has shown remarkable potential for solving challenging segmentation tasks~\cite{zheng_graphcutloss_2021}, for exact or approximate energy minimization in low-level vision with a practical polynomial-time complexity~\cite{boykov_maxflow_2004,jamrivska_gridcut_2012,jensen_mincutreview_2022}. However, the graph-cut method heavily relies on a ``good'' cost function map~\cite{bleyer2005graph} and purely using low-level pixel intensity features~\cite{shi2000normalizedcut} does not give a ``good'' representation of the cost function in complicated image contexts.


With superior data representation learning capacity, deep learning (DL) methods are emerging as a new generation of image segmentation alternatives with remarkably improved performance over traditional image segmentation algorithms~\cite{litjens2017survey,shen2017deep,guo2018review,minaee_dlsegmentationsurvery_2021,liu_dlsegmentationreview_2021,mo2022review}, resulting in a paradigm shift in the field. However, DL segmentation algorithms often need extensive training data~\cite{minaee_dlsegmentationsurvery_2021,tajbakhsh2020embracing}, which poses significant challenges, especially, for medical image segmentation due to patient privacy and high cost. In addition, almost all widely used segmentation models, such as UNet~\cite{ronneberger2015_unet}, FCNs~\cite{long2015fully}, and DeepLab~\cite{chen2017deeplab}, are classification-based in nature and the output probability maps are relatively unstructured, thus lacking the capability of capturing global structures of the target objects. To characterize the long-range data dependency,  transformer~\cite{vaswani_attention_2017,strudel2021segmenter,thisanke2023semantic_visionTransformer} has been introduced for semantic image segmentation, such as TransUNet~\cite{chen2021transunet}, SwinUNet~\cite{cao2022swinunet}, DS-TransUNet~\cite{lin2022ds_transunet}, and nnFormer~\cite{zhou2021nnformer}, which, however, substantially increases the inference cost and memory complexity of the segmentation models. Recent research has demonstrated that, compared to the segmentation CNNs alone, the integration of a graphical model such as conditional random fields (CRFs) into CNNs enhances the robustness of the method to adversarial perturbations~\cite{xie2017adversarial_segmentation,arnab2018robustness,chen2020seqvat}.

Very recently, large data models for image segmentation like SAM~\cite{kirillov2023SAM} and MedSAM~\cite{ma2023MedSAM} have been emerging quickly. However, their high demands on computation resources, e.g. 1600GB GPU memory for training, heavily restrict their applications in general scientific research settings. While a small model, e.g. our proposed model needs only 24GB GPU memory for training, with textbook quality data~\cite{gunasekar2023textbooks} for a specific application is still a major and practical research direction.    


In this paper, we propose to seamlessly integrate the traditional graph-cut and deep learning methods for binary image segmentation,  unifying the strengths of both methods while alleviating the drawbacks of each individual one. Our proposed framework makes use of deep learning networks to learn a high-level feature cost map, and then apply the graph-cut method to achieve a globally optimal segmentation while minimizing the graph-cut energy function defined on the learned cost map. As feature learning and graph-cut optimization are unified in a single deep learning network for end-to-end training, the learned features are tailored specifically for the graph-cut segmentation model with backward propagation.


The challenge in this integration is how to incorporate graph-cut algorithms into deep learning with effective backward propagation support for model training. The combinatorial nature of graph-cut algorithms hampers their applications in deep learning networks due to ineffective backward propagation. Some practical solutions to those combinatorial optimization problems~\cite{smith_nnforco_1999,bengio_mlforco_2021,kotary_e2ecosntrainedoptlearning_2021} are to utilize a good approximation by leveraging the special structures in the problem~\cite{larson_urbanoperationresearch_1981,bengio_mlforco_2021}. With the approximation surrogate, a fundamental quandary persists, that is, the gradient of the optimal solution with respect to the (dynamic) input variables (e.g., the graph edge weights in the graph-cut problem) is frequently zero, which is not helpful for backward propagation to optimize the network parameters~\cite{poganvcic_diffblackboxcombisolver_2019}. Many methods for solving this zero gradient quandary have been explored. Pogan{\v{c}}i{\'c} et al. proposed a differentiation of black box combinatorial solver~\cite{poganvcic_diffblackboxcombisolver_2019}, which, however, doubles the computation burden because of invoking the combinatorial algorithm two times to obtain the gradients. Elmachtoub and Grigas' SPO (Smart “Predict, then Optimize”)~\cite{elmachtoub_smartpredictoptimize_2021} directly leverages the optimization problem structure -- that is, its objective and constraints -- for designing better prediction models. Mensch et al. developed a differentiable dynamic programming method~\cite{mensch_ddp_2018,xie_DDP_2022,xie2023DDP} using a surrogate function for the operator of maximum, which is problem-specific. Khalil et al. proposed learning combinatorial optimization algorithms on
graphs by using neurons to construct graphs dynamically, and heuristically exploring the optimal solution by reinforcement learning mechanism~\cite{khalil_embeddinggraph_2017}. Gasse et al. used graph convolutional neural networks~\cite{gasse_graphcnn_2019} to reformulate branch-and-bound as a Markov decision process for solving a mixed-integer linear programming problem. Both Khalil et al.'s and Gasse et al.'s methods directly used networks to ``simulate" graphs for combinatorial optimization, they yet lost the practical polynomial-time complexity for solving the graph-cut problem~\cite{boykov_maxflow_2004,jamrivska_gridcut_2012,jensen_mincutreview_2022}.            


We proposed a novel residual graph-cut loss and a quasi-residual connection to seamlessly unify a U-Net\cite{ronneberger2015_unet} for feature learning with a graph-cut module to achieve end-to-end training and optimization for binary semantic segmentation. The proposed method is termed as {\em gcDLSeg}. It effectively utilizes the backward propagation from the downstream graph-cut optimization module to guide feature learning, yielding statistically significant improved segmentation accuracy and improved robustness against adversarial attacks. Experiments on public AZH chronic wound~\cite{wang_woundseg_2020} data set and pancreas cancer data set from MSD (medical segmentation decathlon~\cite{antonelli2022_msd}) demonstrated promising segmentation performance. To the best of our knowledge, this is the first work to integrate graph-cut within deep learning for end-to-end training and inference in medical image applications. This proposed method has the potential to be adapted for broader DL with graph-cut applications. 

\section{Methods}
\subsection{Problem Formulation}
Greig et al. ~\cite{greig_mpe_1989} and Boykov et al.~\cite{boykov_maxflow_2004} developed the graph-cut approach to formulate the problem of binary semantic segmentation as a minimum $s$-$t$ (source-sink) cut problem by minimizing an energy function defined on a graph.

\begin{figure}[htbp]
	\includegraphics[width=0.5\textwidth]{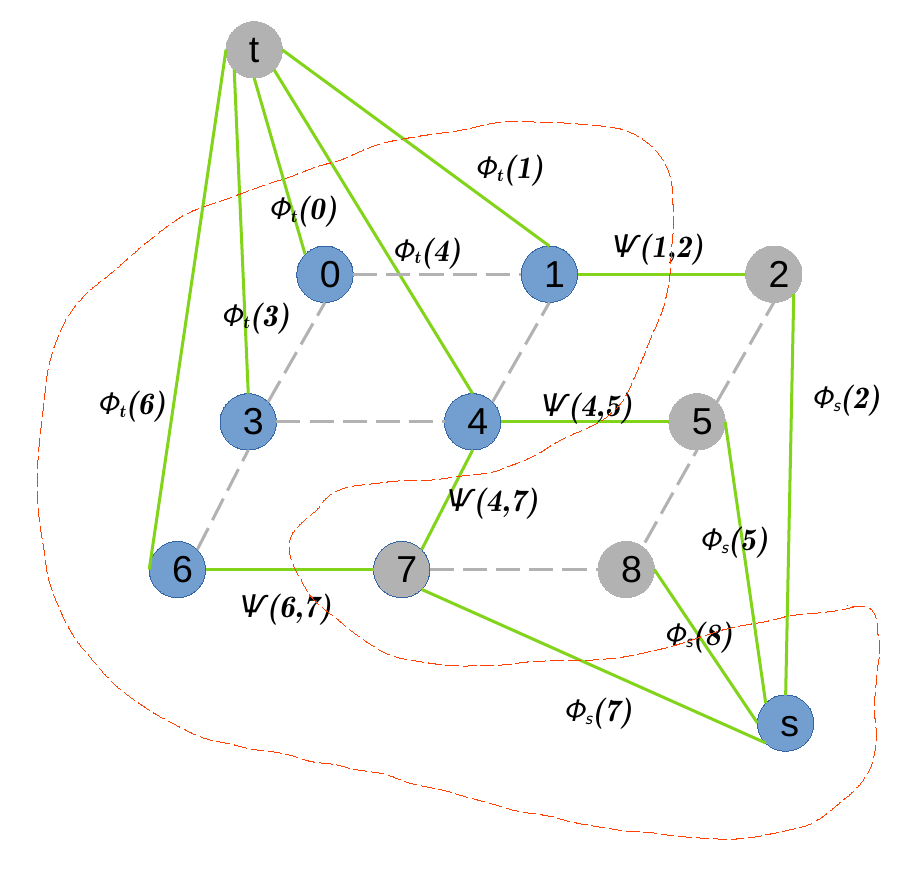}
	\caption {The graph representation of a cut in a 4-neighbor graph system. Number {0,1,2,...,8} indicate pixel nodes $p_i$ in an image $I$ where $i=0,1,2,...,8$ for conciseness, nine pixel nodes are connected by gray dash lines, and $s$ and $t$ are virtual source and sink nodes. A red dash line cuts all nodes into two node sets: source node-set $S$ blue, and sink node-net $T$ gray. Green connections indicate all cut edges between $S$ and $T$.    $\phi_t$ and    $\phi_s$ express t-links, and $\psi$ expresses n-links. Other links that do not participate in current $S-T$ computations are not drawn on the figure for clarity.}
	\label{fig:mincut_graph_new}
\end{figure}
Let $\mathcal{I}$ denote the input image and $\mathcal{N}$ denote its neighboring setting. A labeling $f : \mathcal{I} \mapsto \{0, 1\}$ is a segmentation of the image $\mathcal{I}$, where we should interpret $0$ and $1$ as standing for ``background (bkg)'' and ``foreground (obj)'', respectively.
For each voxel $p\in \mathcal{I}$, we are given a {\em data consistency} function $\alpha_p : \{0,1\} \mapsto \mathbb{R}$, where $\alpha_p(0)$ ($\alpha_p(1)$) represents some pre-computed penalty for assigning voxel $p$ to the background (foreground).  Thus, $\alpha_p(f_p)$ measures the fit of the label $f$ at each voxel $p$ to the foreground ($f_p = 1$) and background ($f_p = 0$).  Similarly, we are given for each voxel pair $(p, q) \in \mathcal{N}$, a {\em pairwise regularization} function $\beta_{pq} : \{0, 1\}\times\{0, 1\} \mapsto \mathbb{R}$, which is the penalty for assigning labels $f_p$ and $f_q$ to two neighboring voxels $p$ and $q$. The purpose is to penalize label differences for any two adjacent voxels. For binary segmentation, the Ising model~\cite{brush1967history_isingmodel} is generally used to design the function $\beta_{pq}$, with $\beta_{pq}(f_p, f_q) =  \psi(p, q)\cdot\delta(f_p=f_q)$, where $\psi(p, q)$ is used to model appearance and smoothness between the pair of voxels $p$ and $q$, and $\delta(\cdot)$ is an identify function giving $1$ if labels $f_p$ and $f_q$ are the same and $0$ otherwise. The segmentation problem seeks to find an optimal labeling $f$ such that the Markov Random Field (MRF) energy function~\cite{komodakis2010mrf} $\mathcal{E}(f)$ is minimized, with
\begin{equation}
	\mathcal{E}(f) = \sum_{p\in\mathcal{I}} \alpha_p(f_p) + \gamma\sum_{(p, q) \in \mathcal{N}} \psi(p, q)\cdot\delta(f_p=f_q),
\end{equation}
\noindent
where $\gamma > 0$ provides the relative weighting between the data consistency terms and the pairwise regularization terms.

The graph-cut segmentation problem can be then readily formulated as a minimum $s$-$t$ cut problem in an edge-weighted graph~\cite{boykov_maxflow_2004}. First, a graph node $p$ represents a voxel in the image $\mathcal{I}$ (to simplify the notation, we use $p$ to denote both a voxel and its corresponding graph node), and edges are added between each pair of nodes $(p, q)$ corresponding to neighborhood voxels with weight $\psi(p, q)$. These edges between neighbors are often referred to as ``$n$-links.''  Next, two extra ``terminal'' nodes, $s$ (the ``obj'' terminal) and $t$ (the ``bkg'' terminal) are added to the graph and for each graph node $p$ corresponding to a voxel in the image, two edges are added (often called ``$t$-links'' for ``terminal''): ($s, p$) and ($p, t$). The weight of each edge between $t$ and node $p$, denoted by $\phi_t(p)$, is given by the background cost value for the associated voxel $p$: $\alpha_p(f_p = `bkg')$, which indicates the coherence between nodes $p$ and $t$. The weight of each edge between node $p$ and ``obj'' terminal $s$, denoted by $\phi_s(p)$, is given by the object cost value for the associated voxel $p$, that is, $\alpha_p(f_p = `obj')$, which indicates the coherence between nodes $p$ and $s$. When determining the minimum $s$-$t$ cut, $(\{s\}\cup S, \{t\}\cup T)$, of this constructed graph, the set $S$ of the nodes remaining connected to the ``obj'' terminal $s$ correspond to the voxels belonging to the segmented object; while the nodes in $T$ correspond to the voxels belonging to the background. Thus, the binary semantic segmentation problem, as illustrated in Fig~\ref{fig:mincut_graph_new}, is formulated to find an $s$-$t$ cut $C = (\{s\}\cup S, \{t\}\cup T)$ in the constructed graph, whose capacity 
\[
||C|| = \sum_{p \in S}\phi_t(p) + \sum_{p \in T}\phi_s(p) + \gamma\sum_{\substack{p \in S\\q \in T\\(p, q) \in \mathcal{N}}}\psi(p, q)
\]
is minimized.

To solve this binary semantic segmentation problem, the proposed {\em gc}DLSeg method first learns the edge weights $\phi_s(p)$ and $\phi_t(p)$ of t-links and $\psi(p, q)$ of n-links in the constructed graph via a U-Net model. Then, a max-flow algorithm~\cite{jamrivska_gridcut_2012} is used to find globally optimal binary labeling for each node $p$ at the current edge weight setting of the graph learned from the network-in-training. The graph edge weight learning and the minimum $s$-$t$ cut computation are implemented in a unified network for an end-to-end gradient backward propagation through a new residual graph-cut loss and a quasi-residual connection for better learning graph edge weights, further improving segmentation accuracy.

\subsection{Network Architecture of {\em gc}DLSeg}

\begin{figure*}[htbp]
	\includegraphics[width=\textwidth]{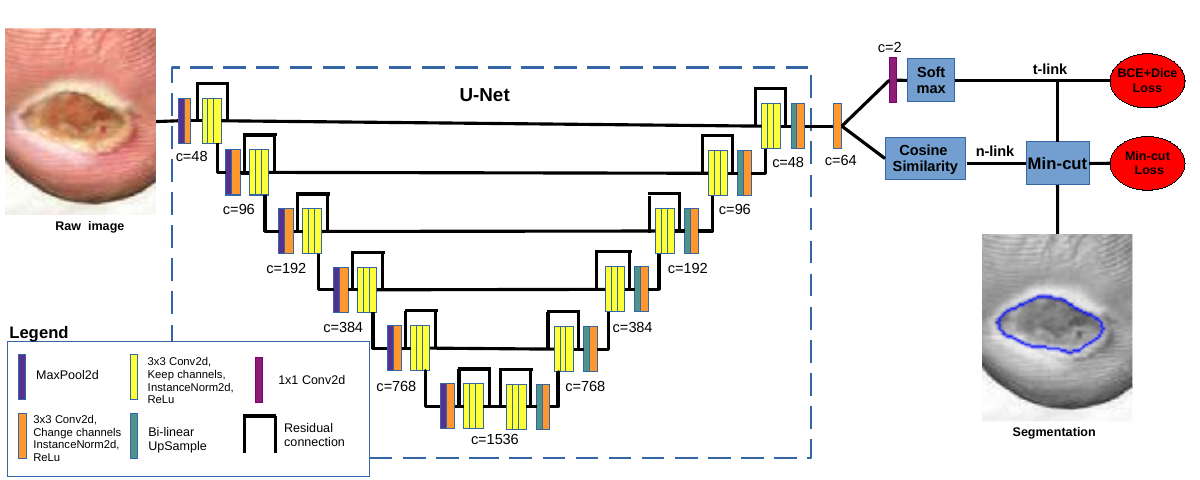}
	\caption{The {\em gc}DLSeg combines U-Net with a graph-cut module for image semantic segmentation. The graph-cut module, with input t-link and n-link edge weights from the preceding feature learning network, solves the max-flow optimization problem for segmentation, outputs optimal segmentation results, and supports the gradient backward propagation of the graph-cut loss signal. $c$ indicates the number of channels for each layer.}
	\label{fig:networkArchitecture_mincut}
\end{figure*}

The proposed semantic segmentation network, {\em gc}DLSeg, is based on a U-Net architecture~\cite{ronneberger2015_unet}, as illustrated in Fig.~\ref{fig:networkArchitecture_mincut}, which consists of six convolution layers. This U-Net acts as a feature-extracting module for the downstream graph-cut segmentation head. The network starts with 48 feature maps in the first convolution layer. In each downsampling layer, a conv2d module followed by a 2x2 max-pooling doubles the feature maps, and then a cascade of three same conv2d modules with a standard residual connection~\cite{hekaiming_residualconnection_2016}. The upsampling layers use one symmetric structure as the downsampling layers but with bilinear upsample modules. 
As we attempt to demonstrate the power of integrating the graph-cut segmentation method with deep learning, the U-Net used in this work was not specifically optimized.  One may also choose any feature-extracting networks to replace the U-Net for specific applications. 

The graph-cut segmentation head in the proposed {\em gc}DLSeg has two branches (Fig.~\ref{fig:networkArchitecture_mincut}). The top t-link branch consists of a 1x1 conv2d and a soft-max module, which serves to learn the two t-link edge weights: $\phi_s(p)$ of the source-to-node edge and $\phi_t(p)$of the node-to-sink edge for each node $p$. We set $\phi_s(p)$ to be the probability of each node $p$ belonging to the source set $S$ and $\phi_t(p)$ to be the probability of $p$ belonging to the sink set $T$, with $\phi_s(p) + \phi_t(p) = 1.0 $.
According to the normalized cut method~\cite{shi2000normalizedcut}, which demonstrated that image segmentation based on low-level cues may not be able to produce a highly accurate segmentation, we compute the similarity of neighbor nodes with abstract high-level features, instead of using the standard Gaussian kernel over RGBXY low-level features~\cite{tang2018normalized,tang2018regularized}. Using the embedding similarity idea~\cite{gibert2012graph_embedding}, the bottom n-link branch in Fig.~\ref{fig:networkArchitecture_mincut} computes the cosine similarity between the feature vectors of two neighboring nodes (voxels), $p$ and $q$, as follows:
\begin{equation}
	\label{eq:consineSimilarityPairWiseItem}
	\psi(p, q) = \frac{1}{2}(1.0 + \frac{<\vec{p}, \vec{q}>}{\| \vec{p}\|    \|\vec{q}\|}),
\end{equation}
where $\vec{p}$ and $\vec{q}$ repersent the feature vectors of voxel nodes $p$ and $q$, respectively.

Finally, the graph-cut module takes the input t-links and n-links to compute the minimum $s$-$t$ cut $C_{min} =(\{s\}\cup S_{min}, \{t\}\cup T_{min})$, in which $S_{min}$ defines the target object of segmentation.

\subsection{Quasi-Residual Connection}
\begin{figure}[htbp]
	\includegraphics[width=0.5\textwidth]{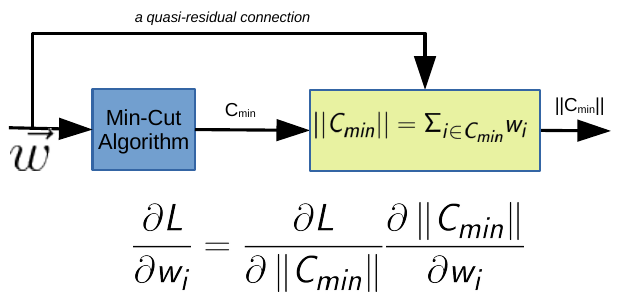}
	\caption{Our proposed {\em gc}DLSeg makes use of the specialty of the graph structure and a quasi-residual connection to
		solve the zero gradient problem in combinatorial algorithms for deep learning. The minimum $s$-$t$ cut algorithm outputs a minimum cut $C_{min}$, which is used in the following cut capacity computation. The $\vec{w}$ is a weight vector of all edges in the graph, $||C_{min}||$ is the min-cut capacity, and $L$ is the loss from its downstream module.}
	\label{fig:proposed_method_solve_zeroGradient}
\end{figure}

We designed a quasi-residual connection, as shown in Fig.~\ref{fig:proposed_method_solve_zeroGradient}, with a residual graph-cut loss to solve the zero gradient quandary in our {\em gc}DLSeg model. The zero gradient quandary is, in some sense, similar to the gradient vanishing problem in RNN~\cite{hochreiter_gradientvanishing_1998}, but it is more difficult to resolve in practice. We consider a residual connection~\cite{hekaiming_residualconnection_2016,residual2016} to provide a pathway for gradients to back-propagate to early layers of the network to avoid vanishing gradient~\cite{ebrahimi_studyonresnet_2021} or zero gradients. The graph-cut module exposes zero gradient to its weights input because of its essence of combinatorial algorithm.
However, our residual connection restores the gradient signal from the downstream loss computation on the same weight input.
We name it ``quasi-residual connection'' as it is not exactly the same as the $y= F(x) +x $ form of the vanilla identity residual connection~\cite{hekaiming_residualconnection_2016,residual2016}. 
The key observation is that the internal of the graph-cut module does not need backward gradient signals as there are no learning parameters within the module.  However, the input (t-links and n-links) to the graph-cut module does need those signals from the downstream loss for feature learning, which is backward propagated via the quasi-residual connection to update the edge weights (Fig.~\ref{fig:proposed_method_solve_zeroGradient}) by bypassing the non-differentiable graph-cut module.

\subsection{Loss Functions}

Directly leveraging the objective and constraints of the optimization problem, the capacity of the output minimum $s$-$t$ cut from the graph-cut module, $C_{min}$, is always less than or equal to the capacity of the {\em ground truth cut}, that is, the $s$-$t$ cut in the constructed graph corresponds to the ground truth segmentation, denote by $C_{gt}$.  We proposed a novel residual graph-cut loss, which essentially converts the pixel-wise classification problem into a regression of the difference between the capacities of $C_{min}$ and $C_{gt}$.

Without loss of generality, assuming the predicted $s$-$t$ minimum cut $C_{min}$ by the graph-cut module is unique, we have 
\begin{equation}
	\begin{aligned}
		||C_{min}|| \leq& \, ||C_{gt}||, \text{where} \\
		||C_{min}|| = &\sum_{p\in S_{min}} \phi_t(p) + \sum_{p\in T_{min}} \phi_s(p)  
		+ \gamma \sum_{\substack{p \in S_{min}\\q \in T_{min}\\(p, q) \in \mathcal{N}}}\psi(p, q) ,\\
		||C_{gt}|| = &\sum_{p\in S_{gt}} \phi_t(p) + \sum_{p\in T_{gt}} \phi_s(p)  
		+ \gamma \sum_{\substack{p \in S_{gt}\\q \in T_{gt}\\(p, q) \in \mathcal{N}}}\psi(p, q) ,
	\end{aligned}
\end{equation}
where $(S_{min}, T_{min})$ is the node partition of the predicted minimum $s$-$t$ cut $C_{min}$, while $(S_{gt}, T_{gt})$ is the node partition of $C_{gt}$. The {\em residual graph-cut loss} is defined, as follows:

\begin{equation}
	\label{eq_Loss_mincut}
		L_{rGC} = \frac{1}{N_o}(||C_{gt}|| -  ||C_{min}||),
\end{equation}
where $N_o$ is a normalization constant which equals the number of graph nodes plus the mean of the numbers of n-links in $C_{min}$ and $C_{gt}$. 
When $L_{rGC} =0$, the predicted min-cut $C_{min}$ equals to the ground truth cut $C_{gt}$; otherwise, $L_{rGC} >0$. 

The proposed $L_{rGC}$ loss unifies two different optimization goals: the graph-cut optimization at the module level, and the ground truth-guided optimization over the whole network level. That is,  $L_{rGC}$ strives to guide the {\em gc}DLSeg network to output a minimum $s$-$t$ cut $C_{min}$ that is the same as the ground truth cut $C_{gt}$. In this way, we convert the pixel-wise segmentation (classification) problem into a regression of the difference between the capacities of $C_{min}$ and $C_{gt}$. \textbf{This regression conversion enables to reformulate the non-differentiable binary node labeling (``obj'' or ``bkg'') in the graph-cut module as a comparison of cut-capacities in the continuous space, which makes the backward propagation feasible.} The graph-cut module has no learning parameters, so the backward gradients can bypass the graph-cut module to be used for the updates of the input t-links and n-links for the module. With the help of the quasi-residual connection, as in Fig.~\ref{fig:proposed_method_solve_zeroGradient}, the backward gradients of $L_{rGC}$ can be backward propagated to early layers of {\em gc}DLSeg and guide the upstream network to learn improved t-link and n-link weights to make a predicted minimum $s$-$t$ cut approaching to the ground truth cut. The effectiveness of the proposed $L_{rGC}$ loss for backward propagation will be proved in Section~\ref{sec-loss-eff}.

In addition to the proposed residual graph-cut loss $L_{rGC}$, we utilize a binary cross entropy loss $L_{ce}$ and a generalized Dice loss $L_{Dice}$~\cite{sudre_gdl_2017}, as follows, to guide the learning of graph edge weights:
\begin{equation}
	\label{eq_Loss_bce}
	L_{ce} = \frac{-\sum_{p}\{ g_p \ln{\phi_s(p)}+ (1-g_p) \ln{\phi_t(p)}\}}{N},    
\end{equation}
where $g_p$ is the ground truth probability that node $p$ belongs to the foreground, and $N$ is the number of total graph nodes.

\begin{equation}
	\label{eq_Loss_dice}
	L_{Dice} = 1.0- 2.0 \frac{w_s\sum_p\{g_p \phi_s(p)\} + w_t\sum_p\{(1-g_p) \phi_t(p)\}}
	{w_s\sum_p\{g_p +\phi_s(p)\} +w_t\sum_p\{(1-g_p) +\phi_t(p)\}},
\end{equation}
where the coefficients $w_s$ and $w_t$ are used to provide invariance to foreground and background set properties, respectively, with $w_s = 1.0/(\sum_p{g_p})^2$ and $w_t = 1.0/(\sum_p{(1-g_p)})^2$.

To find a good set of weights for multiple loss terms in our loss function design, we use the coefficient of variations (relative standard deviation)~\cite{rick_covlossweight_2021} of component losses as the weight for each loss term. Considering the idea that a loss term is satisfied when its relative standard variance is decreased towards zero~\cite{rick_covlossweight_2021}, we define the loss term weight $\alpha^{(i)}_{\lambda}$ for loss term $i$ at time step $\lambda$, as follows. Here the time step $\lambda$ indicates the iterative step for optimizing the network loss. For a sequence $\{b_1, b_2, \ldots, b_{\kappa}\}$, we use $\sigma(b_\kappa)$ and $\mu(b_\kappa)$ to denote the standard deviation and the mean of the sequence. Let $L^{(i)}_{\lambda}$ be the observed value of the $i$-th loss term ($i = 1, 2, 3$) at the time step $\lambda$. We define the loss ratio $r^{(i)}_\lambda = \frac{L^{(i)}_{\lambda}}{\mu(L^{(i)}_{\lambda-1})}$. Then,
%
\begin{equation}
	\begin{aligned}
		\alpha^{(i)}_{\lambda} &= \frac{1}{z_\lambda} \cdot \frac{\sigma\left(r^{(i)}_{\lambda}\right)}{\mu\left(r^{(i)}_{\lambda}\right)},\\
		\text{where } z_\lambda &= \sum_i \frac{\sigma\left(r^{(i)}_{\lambda}\right)}{\mu\left(r^{(i)}_\lambda\right)}.\\    
	\end{aligned}
\end{equation}
Note that $z_\lambda$ is a normalizing constant independent of the number of loss terms to ensure that $\sum_i \alpha^{(i)}_{\lambda} = 1.0$, which is important to decouple the loss term weights from the learning rate~\cite{rick_covlossweight_2021}. The total loss $L_{total}$ at time step $\lambda$ for this binary semantic segmentation network is defined, as follows:
\begin{equation}
	L_{total} = \alpha^{(1)}_{\lambda} L_{ce} + \alpha^{(2)}_{\lambda} L_{Dice} + \alpha^{(3)}_{\lambda} L_{rGC}.
	\label{eqn-total-loss}
\end{equation}

\subsection{Effectiveness of the residual graph-cut loss}
\label{sec-loss-eff}

This section shows the effectiveness of the proposed residual graph-cut loss on the {\em gc}DLSeg model training by backward propagation. We prove that the minimum $s$-$t$ cut capacity is differentiable almost everywhere over the graph edge weights. With the differentiability of the min-cut capacity, the backward gradient of the proposed residual graph-cut loss can be effectively used to update the edge weights in the graph to facilitate the predicted minimum $s$-$t$ cut gradually converging to the ground truth cut during the model training. For the convenience of presentation, in this section, rather than using a node partition $(\{s\}\cup S, \{t\}\cup T)$ to represent an $s$-$t$ cut $C$, we use the set of edges $(u, v)$ with $u\in \{s\}\cup S$ and $v \in  \{t\}\cup T$ to represent $C$. 

\begin{theorem}[\textbf{The derivatives of the min-cut capacity}]
	\label{theorem:Diff_mincut_capacity_2}
	Given an $s$-$t$ graph $G = (\{s, t\}\cup V, E)$ with total $|V|$+2 nodes and $|E|$ edges, let  $\vec{w} \in R^{|E|}$ be the edge weight vector, where each edge weight $w_i \geq 0$, and $C_{min}$ be the set of edges in a minimum $s$-$t$ cut. 
	\\ (1) The derivative of the min-cut capacity $||C_{min}||$ with respect to any edge weight $w_i$ is, as follows:
	\begin{equation}
		\frac{\partial ||C_{min}||}{\partial w_i} =
		\begin{cases}
			1, \quad \forall i \in C_{min},\\
			0, \quad \forall i \not\in C_{min},
		\end{cases}
	\end{equation}
	when the min-cut capacity $||C_{min}||$ is strictly smaller than the cut capacity of any other ones. 
	\\ (2) The min-cut capacity $||C_{min}||$ is differentiable over the vector $\vec{w}$ almost everywhere (except the minimum $s$-$t$ cut $C_{min}$ is not unique) in $R^{|E|}$.    
\end{theorem}

\begin{proof}
	We first consider the case where the min-cut capacity $||C_{min}||$    is strictly less than the cut capacity of any other cuts. Consider    every possible weight $\vec{w}+\vec{\delta}$, where    $\vec{\delta} \in R^{|E|}$    and $\|\vec{\delta}\|_2 \leq \epsilon$ with $\epsilon > 0$ being a small constant. When $\epsilon$ is small enough, under every such weight    $\vec{w}+\vec{\delta}$, the set of original min-cut edges $C_{min}$ under    $\vec{w}$ will remain the set of min-cut edges under $\vec{w}+\vec{\delta}$. This is because under the new weight $\vec{w}+\vec{\delta}$, the cut capacity for $C_{min}$ will at most increase by $\sqrt{|E|} \epsilon$ (by Cauchy-Schwarz inequality), while the cut capacity for any other cut will at most decrease by $\sqrt{|E|} \epsilon$. As long as $2 \sqrt{|E|}\epsilon$ is smaller than the cut-capacity gap between the capacity of any other cut under $\vec{w}$ and the min-cut capacity under $\vec{w}$,    $C_{min}$ will remain the min-cut edge set under $\vec{w}+\vec{\delta}$.

	Then,  the new min-cut capacity under the edge weights $\vec{w}+\vec{\delta}$ is $||C_{min}|| = \sum_{i \in C_{min}} (w_i + \delta_i)$. So we can get the derivative of the min-cut capacity with regard to any edge weight $w_i$, as follows:

	\begin{equation*}
		\frac{\partial ||C_{min}||}{\partial w_i} = \lim_{\delta_i \rightarrow 0} \frac{\Delta ||C_{min}||}{ \delta_{i} }\\
		= \begin{aligned}
			\begin{cases}
				\lim_{\delta_i \rightarrow 0}\frac{\delta_i}{\delta_i}=1, \forall i \in C_{min},\\
				\lim_{\delta_i \rightarrow 0}\frac{0}{\delta_i} =0, \forall i \not\in C_{min},\\
			\end{cases}
		\end{aligned}
	\end{equation*}
	where $\Delta ||C_{min}||$ expresses the change of min-cut capacity under perturbation $\vec{\delta}$.
	
	Now let us consider the case where the min-cut capacity is not strictly less than the capacity of any other cuts.    Suppose that    there exist two different $s$-$t$ cuts $C_{min}^{(1)}$ and $C_{min}^{(2)}$ with the same min-cut capacity, namely $||C_{min}^{(1)}|| = ||C_{min}^{(2)}||$. For some edge $w_i$ where $i \in C_{min}^{(1)} \land i \not\in C_{min}^{(2)}$,    we consider a small perturbation $\vec{\delta}$ such that only $\delta_i$ is non-zero. Then,    we have

	
	\begin{equation*}
		\begin{aligned}
			\begin{cases}
				\lim_{\delta_i \rightarrow 0^+} \frac{\Delta ||C_{min}||}{ \delta_{i} } =\quad \frac{||C_{min}^{(2)}|| - ||C_{min}^{(2)}||}{\delta_i}&=0, \\
				\lim_{\delta_i \rightarrow 0^-}\frac{\Delta ||C_{min}||}{\delta_{i} }=\quad \frac{||C_{min}^{(1)}||+\delta_i - ||C_{min}^{(1)}||}{\delta_i}                     &=1.
			\end{cases}
		\end{aligned}
	\end{equation*}
	
	It shows that at some edge $w_i$ where $i \in C_{min}^{(1)} \land i \not\in C_{min}^{(2)}$, the derivative of the min-cut capacity with respect to $w_i$ is undefined. We thus have the conclusion that the derivatives of the minimum $s$-$t$ cut capacity are undefined for the weights of some edges when multiple cuts achieve the same min-cut capacity (when more than two $s$-$t$ cuts achieve the same min-cut capacity, similar arguments apply by considering an edge belonging to some cuts achieving the min-cut capacity, but not belonging to the other cuts achieving the same min-cut capacity).     
	
	However,    if $||C_{min}^{(1)}|| = ||C_{min}^{(2)}||$, we have $\sum_{i \in C_{min}^{(1)}} w_i = \sum_{i \in C_{min}^{(2)}} w_i$. Because the two sets $C_{min}^{(1)}$ and $C_{min}^{(2)}$ are distinct, this equation defines a hyperplane in the high-dimension space $R^{|E|}$. All hyperplanes have Lebesgue measure zero~\cite{goffman1975_hyperplane_lebesgueMeasure}, and for an absolutely continuous
	distribution, the probability of hitting a set of zero Lebesgue measure is zero~\cite{borovkov2013_lebesgue_zero,rao2016_lebesgue_zero}. (The Lebesgue measure being zero means that $||C_{min}^{(1)}|| = ||C_{min}^{(2)}||$ is an improbable event, such that one can ignore this event for ``most practical purposes'' in the high-dimension space $R^{|E|}$. For example, the probability of choosing a specific value $x$, $0 \leq x    \leq 1$, from a uniform distribution [0,1] is zero, that is, it is an unlikely event. There are at most $2^{|V|}$ cuts for this graph and at most $2^{2|V|}$ such equations (a finite number of hyperplanes) defining two cut capacities being equal. So the set of edge weights such that two cut-capacities are equal is of Lebesgue measure zero.
	
	Therefore, we claim that the min-cut capacity is differentiable over the vector $\vec{w}$ almost everywhere (except when there exist two $s$-$t$ cuts achieving the same min-cut capacity) in $R^{|E|}$. 
\end{proof}

\begin{theorem}[\textbf{The effectiveness of the residual graph-cut loss $L_{rGC}$}]
	The backward gradients of the residual graph-cut loss $L_{rGC}$  facilitate the weight decrease of cut edges unique in the ground truth cut and the weight increase of those edges unique in the predicted minimum $s$-$t$ cut, simultaneously. It promotes the predicted min-cut in the next training iteration to gradually approach the ground truth cut.    	
\end{theorem}

\begin{proof}
	Using the quasi-residual connection, the gradient of the residual graph-cut loss $L_{rGC}$ can be backward propagated to update the graph edge weights in the next training iteration. Recall that the residual graph-cut loss in formula~(\ref{eq_Loss_mincut}), $L_{rGC} = \frac{1}{N_o}(||C_{gt}|| -  ||C_{min}||)$, where $C_{gt}$ is the ground truth cut and $C_{min}$ is the predicted minimum $s$-$t$ cut, and $N_o$ is a normalization constant $N_o > 0$. After canceling the common edges between $C_{gt}$ and $C_{min}$, $L_{rGC}$ can be further expressed, as follows:
	\begin{equation*}
		\begin{aligned}
			L_{rGC} = \frac{1}{N_o} \left( \sum_{i \in C_{gt} \land i \not\in C_{min} }w_i^{(gt)} 
			- \sum_{i \not\in C_{gt} \land i \in C_{min}}w_i^{(min)} \right),
		\end{aligned}
	\end{equation*}
	where, for clearer presentation, $w_i^{(gt)}$ denotes the weight of a cut edge unique in $C_{gt}$ and $w_i^{(min)}$ denotes the weight of a cut edge unique in $C_{min}$.
	
	Using Theorem~\ref{theorem:Diff_mincut_capacity_2} (the derivatives of the min-cut capacity), we can get the backward gradient formulas, as follows:
	\begin{equation*}
		\begin{aligned}
			\frac{\partial L_{rGC}}{w_i^{(gt)}} = \frac{1}{N_o},
			\forall i \in C_{gt} \land i \not\in C_{min}; \\
			\frac{\partial L_{rGC}}{w_i^{(min)}} = -\frac{1}{N_o}, 
			\forall i \not\in C_{gt} \land i \in C_{min}. 
		\end{aligned}
	\end{equation*} 
	
	Using a standard gradient descent method, the edge weights are updated, as follows:
	\begin{equation*}
		\begin{aligned}
			w_i^{(gt)}    &\leftarrow    w_i^{(gt)} - \alpha    \frac{\partial L_{rGC}}{w_i^{(gt)}} \\
			&=    w_i^{(gt)} - \frac{\alpha}{N_o}, 
			\forall i \in C_{gt} \land i \not\in C_{min}, \\
			w_i^{(min)}    &\leftarrow    w_i^{(min)} - \alpha    \frac{\partial L_{rGC}}{w_i^{(min)}} \\
			&=    w_i^{(min)} + \frac{\alpha}{N_o},  \forall i \not\in C_{gt} \land i \in C_{min}
		\end{aligned}
	\end{equation*} 
	where $\alpha > 0 $ is the learning rate during training, and $N_o > 0$. The edge weight update formulas indicate that the backward gradient of the residual graph-cut loss $L_{rGC}$  facilitates the decrease of the weight of cut edges unique in the ground truth cut, and promotes the increase of the weight of cut edges unique in the predicted minimum $s$-$t$ cut, simultaneously. These updated edge weights decrease $||C_{gt}||$ while increasing $||C_{min}||$, which leads to $||C_{min}||$  converging to $||C_{gt}||$. In other words, the predicted minimum $s$-$t$ cut in the next iteration during training gradually approaches the ground truth cut. When $||C_{gt}|| =  ||C_{min}||$, $L_{rGC} =0$, which implies the predicted minimum $s$-$t$ cut overlaps with the ground truth cut. 
		
\end{proof}

\section{Experiments}
The proposed method was validated on two public data sets: the Advancing Zenith of Healthcare (AZH) chronic wound dataset~\cite{wang_woundseg_2020} from the AZH Wound and Vascular Center, Milwaukee, Wisconsin, USA, and the pancreas cancer dataset from the medical segmentation decathlon (MSD)~\cite{antonelli2022_msd}.

The PyTorch version 1.12~\cite{pytorch_2019} on Ubuntu Linux 20.04 was used for the experiments of the proposed method {\em gc}DLSeg. We chose  GridCut~\cite{jamrivska_gridcut_2012} as the implementation of the minimum $s$-$t$ cut algorithm~\cite{jensen_mincutreview_2022}. To evaluate the segmentation performance, we followed the measurements of the compared methods, in which precision, recall, and Dice coefficient were adopted as the evaluation metrics~\cite{minaee_dlsegmentationsurvery_2021}. In order to further demonstrate the performance of our proposed method, we also added the surface-based metrics: average surface distance (ASD), Hausdorff Distance (HD), and 95\%HD. Similar to HD, the 95\%HD is based on the calculation of the 95th percentile of the distances between boundary points in prediction and ground truth, whose purpose is to eliminate the impact of a very small subset of the outliers. For the purpose of the ablation study, we removed the graph-cut module from the proposed {\em gc}DLSeg network and trained the segmentation model with the loss of $L_{ce}$ and $L_{Dice}$. The resulting model is termed as ``NoGraph-Cut.''

The experiments showed that our proposed {\em gc}DLSeg method outperformed the state-of-the-art methods in both metrics of Dice coefficient and recall, and all three surface-based metrics. The proposed {\em gc}DLSeg also demonstrated improved robustness against adversarial attacks.

\subsection{AZH chronic wound segmentation}

The AZH chronic wound data set~\cite{wang_woundseg_2020} consists of 831 training images and 278 test images, each of which is of size $224 \times 224$ pixels with zero-padding. With various backgrounds, the raw images were taken by Canon SX 620 HS digital camera and iPad Pro under uncontrolled illumination conditions~\cite{wang_woundseg_2020}.

In our experiments, we randomly divided the 831 training images into training (706, 85\%) and validation (125, 15\%) sets and kept the original test set (278) untouched. We used 64 channels in the segmentation head, a batch size of 4, and an Adam optimizer with an initial learning rate of 0.0001 without weight decay. We used data augmentation on the fly, including random blurring/sharpening, color space transform, histogram equalization, color cast, white balance, flip, scale, rotation, and translation. 

\begin{figure}[htbp]
	\includegraphics[width=0.5\textwidth]{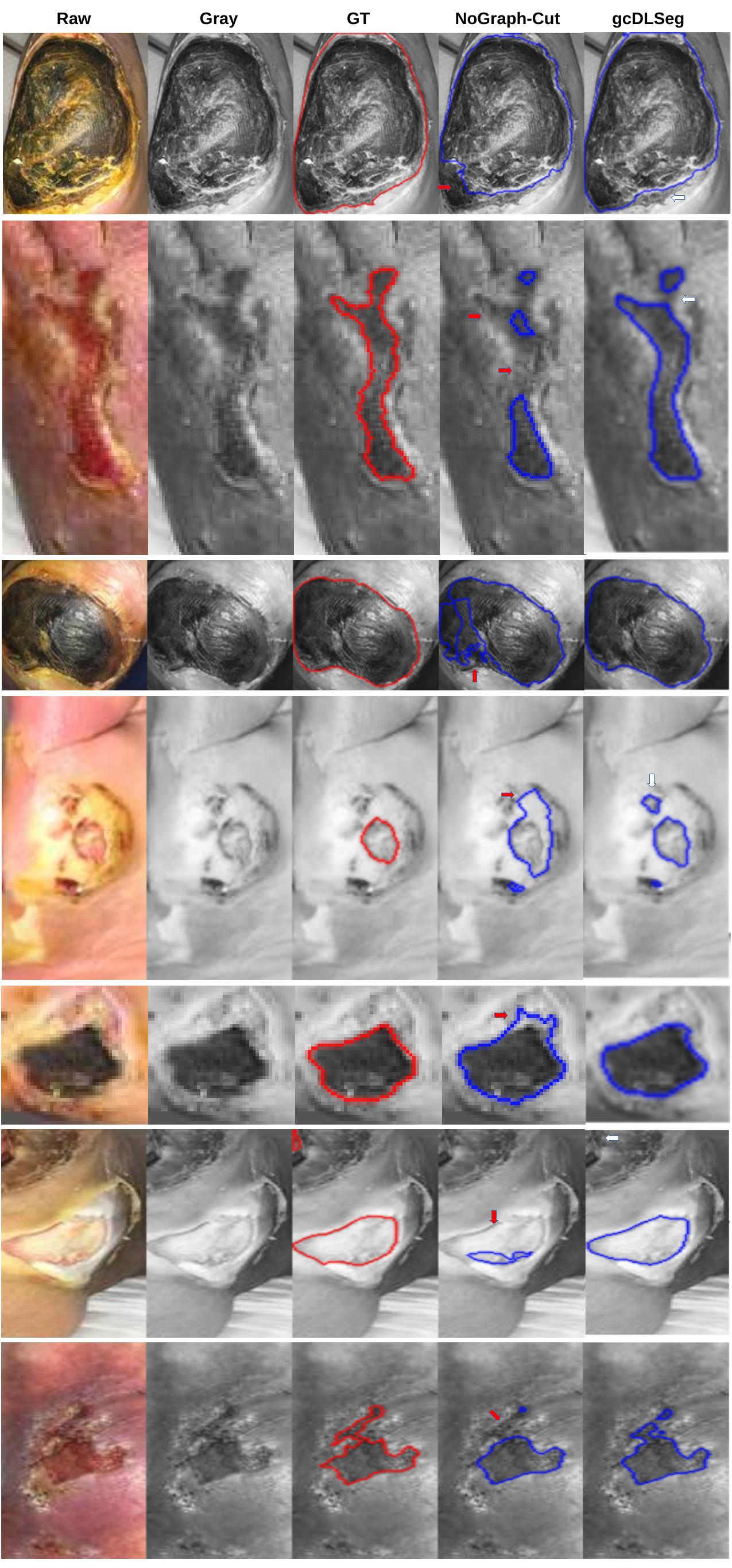}
	\caption{Segmentation samples of seven cases from the AZH chronic wound test dataset. The red arrows indicate segmentation errors. The white arrows show that the results of our proposed {\em gc}DLSeg method are also not perfect.}
	\label{fig_AZH_visualSamples}
\end{figure}

The proposed {\em gc}DLSeg method was compared to various deep learning segmentation models, as presented in Ref.~\cite{wang_woundseg_2020}. In addition, we also compared to Zheng {\em et al.}'s method~\cite{zheng_graphcutloss_2021} termed as ``GraphCutsLoss", in which the graph-cut energy function is used as part of the loss to boost model segmentation accuracy. However, their method does not explicitly integrate the graph-cut segmentation model into the network for model training and inference. Table~\ref{table_AZH_accuracy} shows the performance of all compared methods. The proposed {\em gc}DLSeg method outperformed all compared methods with respect to the metrics of recall and Dice coefficient and ranked second with respect to the metric of precision (Table~\ref{table_AZH_accuracy}). Our proposed {\em gc}DLSeg method also exhibits explicit improvement on the surface position error compared with the NoGraph-Cut method, as in Table~\ref{table_AZH_accuracy_surfaceerror}. The visual segmentation results from the AZH test set are shown in Fig.~\ref{fig_AZH_visualSamples}.

\begin{table}[htbp]
	\caption{Our proposed {\em gc}DLSeg method outperformed all other compared methods in the metrics of recall and Dice coefficient on the AZH test set. The best and second-best results are marked in bold red and black.}
	\label{table_AZH_accuracy}
	\centering
	\resizebox{0.5\textwidth}{!}{%
		\begin{tabular}{|l|c | c | c |}
			\hline
			Methods                                                             										&    Precision (\%)                    			& Recall (\%)                             					& Dice (\%) \\
			\hline
			VGG16\cite{wang_woundseg_2020}                        										& 83.91                                        			& 78.35                                         					& 81.03 \\
			SegNet\cite{wang_woundseg_2020}                     										& 83.66                                        			& 86.49                                        					        & 85.05\\
			U-Net\cite{wang_woundseg_2020}                        										& 89.04                                        			& \textbf{91.29}                                         		        & 90.15\\
			Mask-RCNN\cite{wang_woundseg_2020}                										& \textcolor{red}{\textbf{94.30}}     & 86.40                                         				        & 90.20\\
			MobileNetV2\cite{wang_woundseg_2020}            										& 90.86                                        			& 89.76                                         				        & 90.30\\
			MobileNetV2+CCL\cite{wang_woundseg_2020} 				 						& 91.01                                        			& 89.97                                         				        & 90.47\\
			GraphCutsLoss~\cite{zheng_graphcutloss_2021}                 				                	& 91.45                                        			& 90.24                                         				        & \textbf{90.84}\\
			{\em gc}DLSeg                                                  	                                                        	& \textbf{91.65}                                        &\textcolor{red}{\textbf{91.99}}                                &\textcolor{red}{\textbf{91.82}}\\     
			\hline
			\multicolumn{4}{l}{Notes: in medical disease measurement, recall is far more important than precision~\cite{powers2020evaluation}.}

		\end{tabular}
	}
	
\end{table}

\begin{table}[htbp]
	
	\caption{Our proposed {\em gc}DLSeg method achieved reduced boundary errors, compared with the GraphcutLoss method, in all three surface-based metrics on the AZH test set. The smaller the value, the better.}
	\label{table_AZH_accuracy_surfaceerror}
	\centering
	\resizebox{0.5\textwidth}{!}{%
		\begin{tabular}{|l|c | c | c |}
			\hline
			Methods        &  ASD (pixel) &	   HD (pixel) &	 95\%HD (pixel) \\
			\hline
			GraphcutLoss   &1.38&	7.76&	4.88\\
			gcDLSeg	       & \textcolor{red}{\textbf{1.30}}	     &  \textcolor{red}{\textbf{6.56}}	&\textcolor{red}{\textbf{4.00}}\\
			\hline
			
		\end{tabular}
	}
	
\end{table}

\begin{table}[htbp]
	\caption{Our proposed {\em gc}DLSeg model shows improved robustness against adversarial attacks. The precision, recall, and Dice coefficient were computed for the proposed {\em gc}DLSeg and the  NoGraph-Cut methods applied on the AZH test set at different adversarial attack scales ($\epsilon$).}
	\label{table_AZH_accuracy_adversarial}
	\centering
	\resizebox{0.5\textwidth}{!}{%
		\begin{tabular}{|c|c    c    c | c    c    c |}
			\hline
			Epsilon         &    \multicolumn{3}{c|}{NoGraph-Cut}     &    \multicolumn{3}{c|}{{\em gc}DLSeg}        \\
			($\epsilon$)& Precision (\%) & Recall(\%)& Dice (\%)        & Precision (\%)    & Recall (\%) &    Dice (\%)    \\
			\hline
			0.00            &    90.11             &    90.51        &    90.31     & 91.65                     & 91.99         & 91.82    \\
			0.02            &    75.20             &    80.28        &    77.65     & 77.20                     & 83.75         & 80.34    \\
			0.04            &    64.69             &    78.45        &    70.91     & 72.39                     & 82.52         & 77.13    \\
			0.06            &    54.54             &    79.30        &    64.63     & 67.11                     & 83.34         & 74.35    \\
			0.08            &    47.77             &    79.44        &    59.66     & 62.16                     & 83.06         & 71.11    \\
			0.10            &    42.46             &    78.39        &    55.08     & 58.40                     & 83.00         & 68.56    \\
			\hline
		\end{tabular}
	}
	
\end{table}

\begin{figure}[htbp]
	\includegraphics[width=0.5\textwidth]{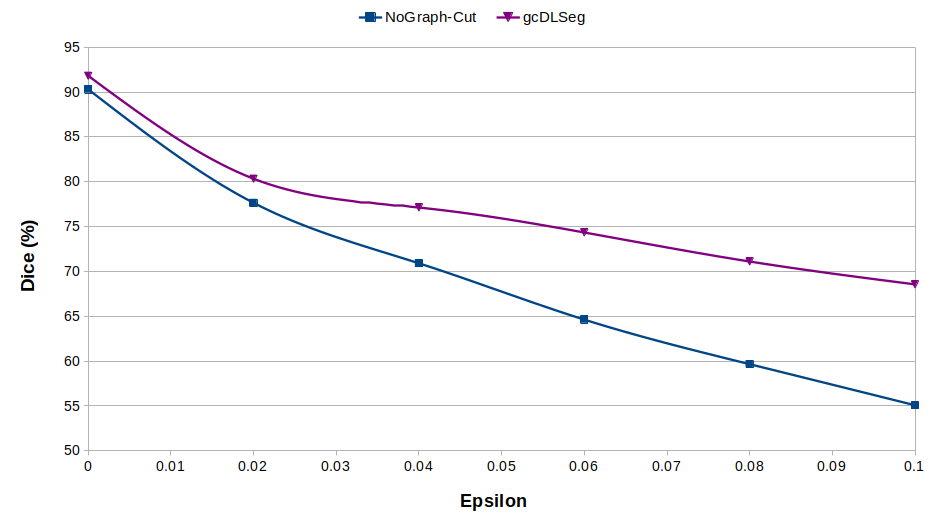}
	\caption{The robustness performance of the proposed {\em gc}DLSeg method against adversarial attacks on the AZH test set. All Dice coefficients decrease with bigger-scale ($\epsilon$) adversarial attacks. However, our proposed method shows a smaller decreasing slope.}
	\label{fig_AZH_adversarialAttack_scale}
\end{figure}

\begin{figure}[htbp]
	\includegraphics[width=0.5\textwidth]{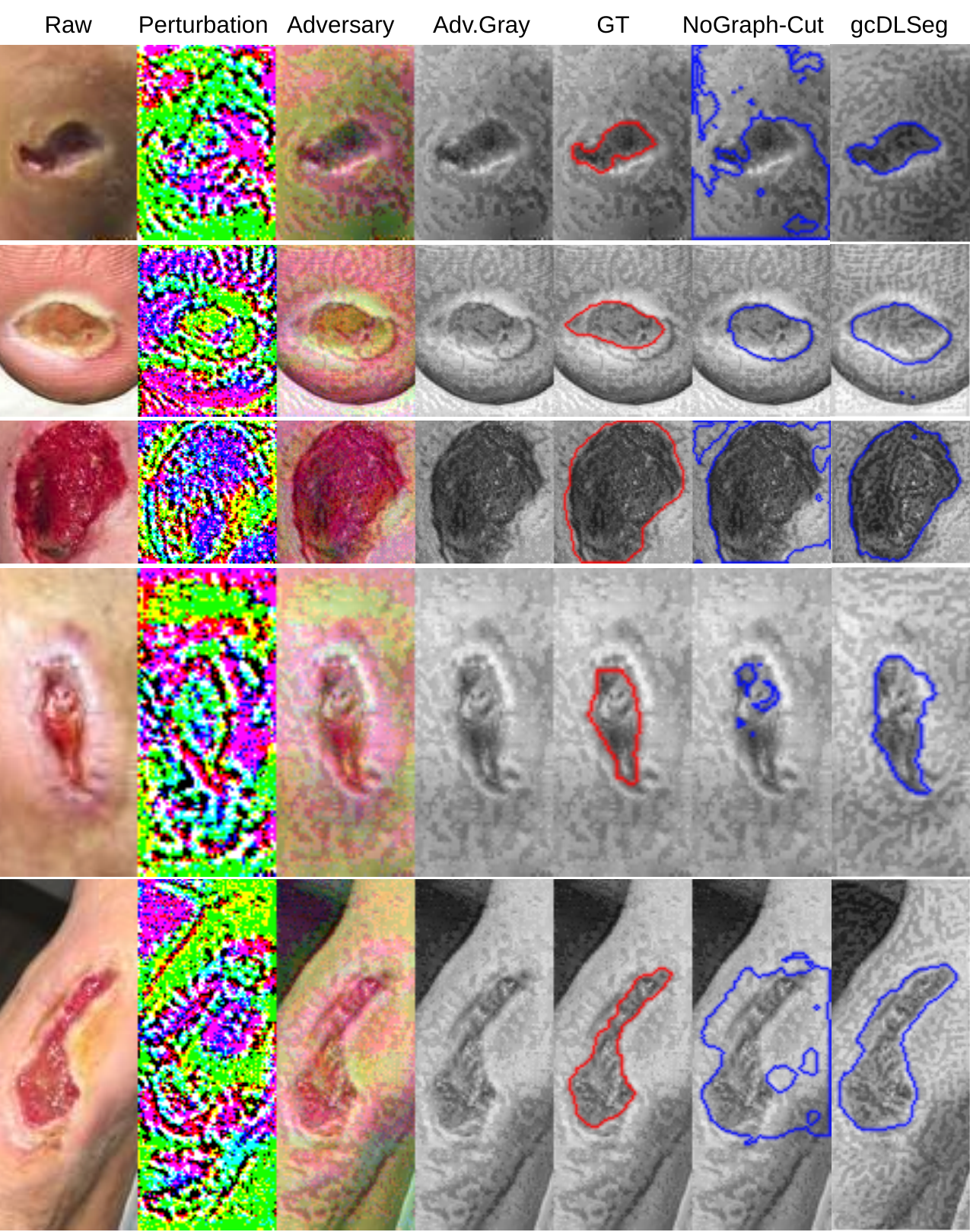}
	\caption{Segmentation samples of five cases from the AZH test set using untargeted white-box adversarial attacks with $\epsilon=0.1$. The adversarial samples (the 3rd column) were generated by adding  0.01 $\cdot$ perturbation images (the 2nd column) to the original images (the first column) for the NoGraph-Cut method. For the proposed {\em gc}DLSeg method, we used its own training loss $L_{total}$ (Eqn.~(\ref{eqn-total-loss})) to generate the adversarial perturbations to maximize its attacks.}
	
	\label{fig_AZH_adversarialSamples}
\end{figure}

Our {\em gc}DLSeg model network also demonstrated improved robustness against adversarial attacks. We used untargeted white-box adversarial attacks to test this capability. Goodfellow {\em et al.}'s fast gradient sign method~\cite{goodfellow_adversarialattack_2014} was used to generate adversarial noise, as follows:
\begin{equation} \label{eq_GoodFellow_FGSM}
	I_{adv} = I + \epsilon \cdot sign(\frac{dL}{dI}),
\end{equation}
where $I$ is the original normalized input images, $I_{adv}$ is the generated adversarial samples, and $L$ is the network loss. The perturbation scale $\epsilon$ changes from $0.0$ to $0.1$ with step size $0.02$. The precision, recall, and Dice coefficient evaluated on the AZH test set using different methods at different adversarial perturbations ($\epsilon$) are shown in Table~\ref{table_AZH_accuracy_adversarial}, and the dice decreasing curve is illustrated in Fig.~\ref{fig_AZH_adversarialAttack_scale}. Their visual results are illustrated in Fig.~\ref{fig_AZH_adversarialSamples}. 

\subsection{Pancreas tumor segmentation}

\begin{figure*}[htbp]
	\includegraphics[width=\textwidth]{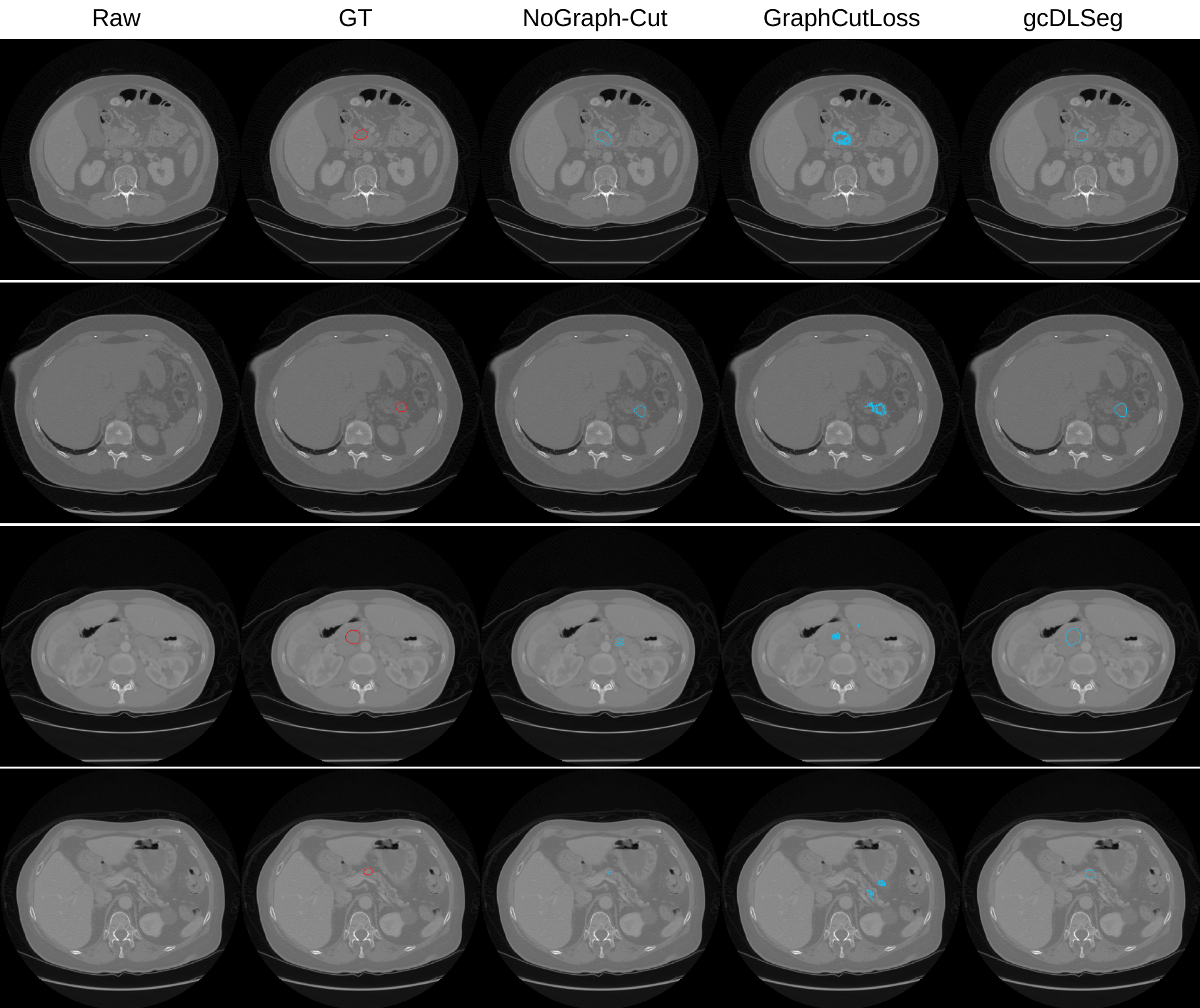}
	\caption{Segmentation samples of four cases in the MSD pancreas cancer test dataset. The tumor boundary in both GT and predictions are drawn by a one-pixel boundary line. The GraphcutLoss method has explicit un-smooth boundary phenomena, even it is drawn by the same one-pixel boundary. Please use the PDF magnifying glass tool to get a better view effect. We keep the full slice view, instead of the view of a region of interest (ROI), in order to show readers that segmenting tiny tumors in a messy abdominal background is very challenging.}
	\label{fig_pancreasCancer_visualSamples}
\end{figure*}

The pancreas cancer data set from the medical segmentation decathlon (MSD)~\cite{antonelli2022_msd} consists of 281 3D volumes of abdominal CT images each with $512 \times 512$ pixels per slice and its corresponding ground truth labels. Segmenting pancreas cancer is the most challenging segmentation task among various MSD data sets. Fig~\ref{fig_pancreasCancer_visualSamples} shows some challenging examples visually. We randomly divided the 281 volumes into three data sets (training, validation, and test) and deleted all slices without the pancreas region. Then we got the training set (1304 slices), the validation set (597 slices), and the test set (634 slices). We used 2D slices as input to our networks and predicted the 2D segmentation for each slice.

We used 64 channels in the segmentation head, a batch size of 4, and an Adam optimizer with an initial learning rate of 0.0001 without weight decay. Data augmentation on the fly was used, including random blurring/sharpening, slight rotation ($[-29^\circ, 29^\circ]$), salt and pepper noise, and speckle noise. 

Among all comparison methods, our proposed {\em gc}DLSeg method improved more than $3\%$ in Dice coefficient over the previous state-of-the-art nnU-Net method, as shown in Table~\ref{table_pancreascancer_accuracy}. Our proposed {\em gc}DLSeg method also exhibits explicit improvement on the surface position error compared with the NoGraph-cut and GraphcutLoss methods, as in Table~\ref{table_pancreascancer_accuracy_surfaceerror}. Their visual example segmentations from the pancreas test set are demonstrated in Fig~\ref{fig_pancreasCancer_visualSamples}.

\begin{table}[htbp]
	\caption{The proposed {\em gc}DLSeg method outperformed all other compared methods in all metrics on the pancreas cancer data set from MSD. The best results are marked in bold red. Blanks mean that the original literature didn't report the corresponding measurements.}
	\label{table_pancreascancer_accuracy}
	\resizebox{0.5\textwidth}{!}{%
		\begin{tabular}{|l| c | c | c|}
			\hline
			Methods                                                                                    &Precision (\%)&Recall (\%)& Dice (\%) \\
			\hline 
			nnU-Net 3D~\cite{isensee2021_nnunet}$^*$          &                        &                                                                                                                &52.00 \\
			K.A.V.athlon 3D $^{**}$                                     &                        &                                                                                                                                                &         43.00 \\
			nnU-Net 2D~\cite{isensee2021_nnunet}                        &                            &                    &35.01 \\
			
			GraphCutLoss~\cite{zheng_graphcutloss_2021}                                                                                                                    &                40.74 &         39.52 &         40.12 \\
			{\em gc}DLSeg                                                           &\textcolor{red}{\textbf{52.78}} &\textcolor{red}{\textbf{57.59}} &\textcolor{red}{\textbf{55.08}} \\
			\hline

		\end{tabular}
	}
	
	\vspace*{0.5mm}
	\tiny{$^*$: Top 1 method in MSD Grand Challenge 2018} \\
	\tiny{$^{**}$: Top 2 method in MSD Grand Challenge 2018}
\end{table}

\begin{table}[htbp]
	
	\caption{Our proposed {\em gc}DLSeg method achieved reduced boundary errors, comparing with GraphcutLoss methods, in all three surface-based metrics on the pancreas cancer data set from MSD. The smaller the value, the better.}
	\label{table_pancreascancer_accuracy_surfaceerror}
	\centering
	\resizebox{0.5\textwidth}{!}{%
		\begin{tabular}{|l|c | c | c |}
			\hline
			Methods        &  ASD (pixel) &	   HD (pixel) &	 95\%HD (pixel) \\
			\hline
			GraphCutLoss   &	16.78	& 33.12 &	29.03\\
			gcDLSeg	        & \textcolor{red}{\textbf{12.30}}	&  \textcolor{red}{\textbf{23.21}}	&  \textcolor{red}{\textbf{21.96}}\\
			\hline
			
		\end{tabular}
	}
	
\end{table}

\subsection{Ablation study}

For the purpose of the ablation study, the NoGraph-Cut model by removing the graph-cut module from the proposed {\em gc}DLSeg network architecture was trained with the loss of $L_{ce}$ and $L_{Dice}$. We also applied the graph-cut segmentation method as post-processing on the probability map output from the NoGraph-Cut model to investigate its performance. The ablation experiments were conducted on both the chronic wound~\cite{wang_woundseg_2020} and the pancreas cancer~\cite{antonelli2022_msd}) data sets. 

The performance of all three methods on both datasets is shown in Table~\ref{table_AZH_accuracy_ablation} and Table~\ref{table_pancreascancer_accuracy_ablation}, respectively. The segmentation results demonstrated that the proposed graph-cut module supported by backward propagation with the residual graph-cut loss within the deep learning network was able to significantly improve segmentation performance with respect to all metrics used, compared to the other two methods.  For the method using graph-cut as postprocessing, the segmentation performance was comparable to its baseline method of NoGraph-Cut.  In this scheme, feature learning by the NoGraph-Cut network is, in fact, disconnected from the graph-cut model; the learned features thus may not be truly appropriated for the graph-cut model. However, in the framework of our proposed {\em gc}DLSeg method, the graph-cut model is used to guide the feature learning directly with U-Net. 
We also performed experiments, which demonstrated that the graph-cut module with backward propagation in fine-tuning training can further improve probability maps. After pre-training of U-Net, the graph-cut module was added for network fine-tuning. The binary cross-entropy loss, which measures the distance between the predicted and ground truth probability distributions, was further reduced. It implies that the backward propagation of the residual graph-cut loss further improved the probability map.

\begin{table}[htbp]
	\caption{Ablation experiments showed the proposed graph-cut module integrated with U-Net for end-to-end training improved segmentation performance on the AZH test set. The best results are marked in bold red.}
	\label{table_AZH_accuracy_ablation}
	\centering
	\resizebox{0.5\textwidth}{!}{%
		\begin{tabular}{|l|c | c | c |}
			\hline
			Ablation Experiments                                                             										&    Precision (\%)                    			& Recall (\%)                             					& Dice (\%) \\
			\hline
			NoGraph-Cut       													& 90.11                                        			& 90.51                                         				        & 90.31\\
			Graph-cut as post-processing                                         			& 91.23                                        			& 89.70                                         				        & 90.46 \\
			
			{\em gc}DLSeg                                             		& \textcolor{red}{\textbf{91.65}}                        		                	&\textcolor{red}{\textbf{91.99}}                                &\textcolor{red}{\textbf{91.82}}\\     
			\hline
			
		\end{tabular}
	}

\end{table}

\begin{table}[htbp]
	\caption{Ablation experiments showed the proposed graph-cut module integrated with U-Net for end-to-end training improved segmentation performance on pancreas cancer data set from MSD. The best results are marked in bold red.}
	\label{table_pancreascancer_accuracy_ablation}
	\centering
	\resizebox{0.5\textwidth}{!}{%
		\begin{tabular}{|l|c | c | c|}
			\hline
			Ablation Experiments                                                                                     &Precision (\%)&Recall (\%)& Dice (\%) \\
			\hline 
			
			NoGraph-Cut                                                                                                                 &                45.53 &         52.32 &         48.69 \\
			Graph-cut as post-processing                                                                         &                46.08 &         51.10 &         48.46 \\
			{\em gc}DLSeg                                                         &\textcolor{red}{\textbf{52.78}} &\textcolor{red}{\textbf{57.59}} &\textcolor{red}{\textbf{55.08}} \\
			\hline

		\end{tabular}
	}
	
\end{table}

\begin{table}[htbp]
	
	\caption{
		Ablation experiments showed the proposed graph-cut module integrated with U-Net for end-to-end training reduced boundary error on the AZH test set. The smaller the value, the better.}
	\label{table_AZH_accuracy_surfaceerror_ablation}
	\centering
	\resizebox{0.5\textwidth}{!}{%
		\begin{tabular}{|l|c | c | c |}
			\hline
			Methods        &  ASD (pixel) &	   HD (pixel) &	 95\%HD (pixel) \\
			\hline
			NoGraph-Cut    & 1.44      &  7.98  &5.20\\
			Graph-cut as post-processing &1.39 &7.76  &5.14 \\
			gcDLSeg	       & \textcolor{red}{\textbf{1.30}}	     &  \textcolor{red}{\textbf{6.56}}	&\textcolor{red}{\textbf{4.00}}\\
			\hline
			
		\end{tabular}
	}
	
\end{table}

\begin{table}[htbp]
	
	\caption{
		Ablation experiments showed the proposed graph-cut module integrated with U-Net for end-to-end training reduced boundary error on the pancreas cancer data set from MSD. The smaller the value, the better.}
	\label{table_pancreascancer_accuracy_surfaceerror_ablation}
	\centering
	\resizebox{0.5\textwidth}{!}{%
		\begin{tabular}{|l|c | c | c |}
			\hline
			Methods        &  ASD (pixel) &	   HD (pixel) &	 95\%HD (pixel) \\
			\hline
			NoGraph-Cut	   & 14.71 & 25.54 & 24.00\\
			Graph-cut as post-processing &14.46	&27.20	&25.76 \\
			gcDLSeg	        & \textcolor{red}{\textbf{12.30}}	&  \textcolor{red}{\textbf{23.21}}	&  \textcolor{red}{\textbf{21.96}}\\
			\hline
			
		\end{tabular}
	}
	
\end{table}

\section{Discussion and Conclusion}

In this study, we developed a novel DL framework for binary semantic image segmentation, which leverages a new residual graph-cut loss and a quasi-residual connection to seamlessly integrate the graph-cut segmentation model with the U-Net segmentation network for end-to-end learning. The residual graph-cut loss essentially enables reformulating the pixel-wise classification problem as a regression problem for capturing the difference between the capacities of the min-cut and the ground truth cut. We theoretically proved the derivativity of the min-cut capacity and the effectiveness of the proposed residual graph-cut loss for feature learning via backward propagation. The quasi-residual connection provides a pathway bypassing the combinatorial and non-differentiable graph-cut optimization module for the backward propagation of gradients to earlier layers of {\em gc}DLSeg. The global optimality ensured by the minimum $s$-$t$ algorithm facilitates better feature learning efficiently during network training. In the inference phase, globally optimal segmentation is achieved with respect to the graph-cut segmentation model defined on the optimized image features from {\em gc}DLSeg.

The major drawback of our proposed method is its training efficiency.  During the network training, it is computationally intensive to run the minimum $s$-$t$ cut algorithm to compute the optimal solutions with large training datasets and training epochs.  
The minimum $s$-$t$ cut algorithm used in our current implementation is GridCut~\cite{jamrivska_gridcut_2012}, which is an augmenting path algorithm and is challenging to parallelize in GPU. Further improvement includes using a parallel push-relabel min-cut algorithm implemented in GPU to improve training and inference efficiency. Another way is, during the early stages of training, to compute approximations to the minimum $s$-$t$ cuts, and then refine the network with optimal $s$-$t$ cuts in the final stages of training.

The proposed method was validated on the public AZH chronic wound data set~\cite{wang_woundseg_2020} and the pancreas cancer data set from MSD~\cite{antonelli2022_msd}). Our experiments showed that the proposed {\em gc}DLSeg method outperformed the state-of-the-art methods in Dice, recall, and surface position error. Our proposed method also demonstrated improved robustness against adversarial attacks. We expect the developed {\em gc}DLSeg method would find broader applications in computer vision, which are involved in the minimum $s$-$t$ cut algorithm. The techniques of integrating graph-cut into the deep learning networks can also be extendable for other combinatorial optimization methods.

\section*{Acknowledgments} Thanks to Patrick M Jensen at the Technical University of Denmark, who suggested using GridCut~\cite{jamrivska_gridcut_2012} as the implementation of the minimum $s$-$t$ cut algorithm. This study was supported in part by NSF grants CCF-1733742 and ECCS-2000425 and in part by NIH grants 1U54HL165442 and 5R01AG067078.

\bibliographystyle{IEEEtran}
\bibliography{ref_AllChapters}

\begin{thebibliography}{10}
\providecommand{\url}[1]{#1}
\csname url@samestyle\endcsname
\providecommand{\newblock}{\relax}
\providecommand{\bibinfo}[2]{#2}
\providecommand{\BIBentrySTDinterwordspacing}{\spaceskip=0pt\relax}
\providecommand{\BIBentryALTinterwordstretchfactor}{4}
\providecommand{\BIBentryALTinterwordspacing}{\spaceskip=\fontdimen2\font plus
\BIBentryALTinterwordstretchfactor\fontdimen3\font minus \fontdimen4\font\relax}
\providecommand{\BIBforeignlanguage}[2]{{%
\expandafter\ifx\csname l@#1\endcsname\relax
\typeout{** WARNING: IEEEtran.bst: No hyphenation pattern has been}%
\typeout{** loaded for the language `#1'. Using the pattern for}%
\typeout{** the default language instead.}%
\else
\language=\csname l@#1\endcsname
\fi
#2}}
\providecommand{\BIBdecl}{\relax}
\BIBdecl

\bibitem{minaee_dlsegmentationsurvery_2021}
S.~Minaee, Y.~Y. Boykov, F.~Porikli, A.~J. Plaza, N.~Kehtarnavaz, and D.~Terzopoulos, ``Image segmentation using deep learning: A survey,'' \emph{IEEE transactions on pattern analysis and machine intelligence}, 2021.

\bibitem{boykov_graphcut_2001}
Y.~Y. Boykov and M.-P. Jolly, ``Interactive graph cuts for optimal boundary \& region segmentation of objects in nd images,'' in \emph{Proceedings eighth IEEE international conference on computer vision. ICCV 2001}, vol.~1.\hskip 1em plus 0.5em minus 0.4em\relax IEEE, 2001, pp. 105--112.

\bibitem{boykov_graphcut_ieee_2001}
Y.~Boykov, O.~Veksler, and R.~Zabih, ``Fast approximate energy minimization via graph cuts,'' \emph{IEEE Transactions on pattern analysis and machine intelligence}, vol.~23, no.~11, pp. 1222--1239, 2001.

\bibitem{boykov_maxflow_2004}
Y.~Boykov and V.~Kolmogorov, ``An experimental comparison of min-cut/max-flow algorithms for energy minimization in vision,'' \emph{IEEE transactions on pattern analysis and machine intelligence}, vol.~26, no.~9, pp. 1124--1137, 2004.

\bibitem{vese2002multiphase}
L.~A. Vese and T.~F. Chan, ``A multiphase level set framework for image segmentation using the mumford and shah model,'' \emph{International journal of computer vision}, vol.~50, pp. 271--293, 2002.

\bibitem{li2010distance}
C.~Li, C.~Xu, C.~Gui, and M.~D. Fox, ``Distance regularized level set evolution and its application to image segmentation,'' \emph{IEEE transactions on image processing}, vol.~19, no.~12, pp. 3243--3254, 2010.

\bibitem{li2011level}
C.~Li, R.~Huang, Z.~Ding, J.~C. Gatenby, D.~N. Metaxas, and J.~C. Gore, ``A level set method for image segmentation in the presence of intensity inhomogeneities with application to mri,'' \emph{IEEE transactions on image processing}, vol.~20, no.~7, pp. 2007--2016, 2011.

\bibitem{greig_mpe_1989}
D.~M. Greig, B.~T. Porteous, and A.~H. Seheult, ``Exact maximum a posteriori estimation for binary images,'' \emph{Journal of the Royal Statistical Society: Series B (Methodological)}, vol.~51, no.~2, pp. 271--279, 1989.

\bibitem{jensen_mincutreview_2022}
P.~M. Jensen, N.~Jeppesen, A.~B. Dahl, and V.~A. Dahl, ``Review of serial and parallel min-cut/max-flow algorithms for computer vision,'' \emph{arXiv preprint arXiv:2202.00418}, 2022.

\bibitem{zheng_graphcutloss_2021}
Z.~Zheng, M.~Oda, and K.~Mori, ``Graph cuts loss to boost model accuracy and generalizability for medical image segmentation,'' in \emph{Proceedings of the IEEE/CVF International Conference on Computer Vision}, 2021, pp. 3304--3313.

\bibitem{jamrivska_gridcut_2012}
O.~Jamri{\v{s}}ka, D.~S{\`y}kora, and A.~Hornung, ``Cache-efficient graph cuts on structured grids,'' in \emph{2012 IEEE Conference on Computer Vision and Pattern Recognition}.\hskip 1em plus 0.5em minus 0.4em\relax IEEE, 2012, pp. 3673--3680.

\bibitem{bleyer2005graph}
M.~Bleyer and M.~Gelautz, ``Graph-based surface reconstruction from stereo pairs using image segmentation,'' in \emph{Videometrics VIII}, vol. 5665.\hskip 1em plus 0.5em minus 0.4em\relax SPIE, 2005, pp. 288--299.

\bibitem{shi2000normalizedcut}
J.~Shi and J.~Malik, ``Normalized cuts and image segmentation,'' \emph{IEEE Transactions on pattern analysis and machine intelligence}, vol.~22, no.~8, pp. 888--905, 2000.

\bibitem{litjens2017survey}
G.~Litjens, T.~Kooi, B.~E. Bejnordi, A.~A.~A. Setio, F.~Ciompi, M.~Ghafoorian, J.~A. Van Der~Laak, B.~Van~Ginneken, and C.~I. S{\'a}nchez, ``A survey on deep learning in medical image analysis,'' \emph{Medical image analysis}, vol.~42, pp. 60--88, 2017.

\bibitem{shen2017deep}
D.~Shen, G.~Wu, and H.-I. Suk, ``Deep learning in medical image analysis,'' \emph{Annual review of biomedical engineering}, vol.~19, pp. 221--248, 2017.

\bibitem{guo2018review}
Y.~Guo, Y.~Liu, T.~Georgiou, and M.~S. Lew, ``A review of semantic segmentation using deep neural networks,'' \emph{International journal of multimedia information retrieval}, vol.~7, no.~2, pp. 87--93, 2018.

\bibitem{liu_dlsegmentationreview_2021}
X.~Liu, L.~Song, S.~Liu, and Y.~Zhang, ``A review of deep-learning-based medical image segmentation methods,'' \emph{Sustainability}, vol.~13, no.~3, p. 1224, 2021.

\bibitem{mo2022review}
Y.~Mo, Y.~Wu, X.~Yang, F.~Liu, and Y.~Liao, ``Review the state-of-the-art technologies of semantic segmentation based on deep learning,'' \emph{Neurocomputing}, vol. 493, pp. 626--646, 2022.

\bibitem{tajbakhsh2020embracing}
N.~Tajbakhsh, L.~Jeyaseelan, Q.~Li, J.~N. Chiang, Z.~Wu, and X.~Ding, ``Embracing imperfect datasets: A review of deep learning solutions for medical image segmentation,'' \emph{Medical Image Analysis}, vol.~63, p. 101693, 2020.

\bibitem{ronneberger2015_unet}
O.~Ronneberger, P.~Fischer, and T.~Brox, ``{U-Net}: Convolutional networks for biomedical image segmentation,'' in \emph{International Conference on Medical image computing and computer-assisted intervention}.\hskip 1em plus 0.5em minus 0.4em\relax Springer, 2015, pp. 234--241.

\bibitem{long2015fully}
J.~Long, E.~Shelhamer, and T.~Darrell, ``Fully convolutional networks for semantic segmentation,'' in \emph{Proceedings of the IEEE conference on computer vision and pattern recognition}, 2015, pp. 3431--3440.

\bibitem{chen2017deeplab}
L.-C. Chen, G.~Papandreou, I.~Kokkinos, K.~Murphy, and A.~L. Yuille, ``Deeplab: Semantic image segmentation with deep convolutional nets, atrous convolution, and fully connected crfs,'' \emph{IEEE transactions on pattern analysis and machine intelligence}, vol.~40, no.~4, pp. 834--848, 2017.

\bibitem{vaswani_attention_2017}
A.~Vaswani, N.~Shazeer, N.~Parmar, J.~Uszkoreit, L.~Jones, A.~N. Gomez, {\L}.~Kaiser, and I.~Polosukhin, ``Attention is all you need,'' in \emph{Advances in neural information processing systems}, 2017, pp. 5998--6008.

\bibitem{strudel2021segmenter}
R.~Strudel, R.~Garcia, I.~Laptev, and C.~Schmid, ``Segmenter: Transformer for semantic segmentation,'' in \emph{Proceedings of the IEEE/CVF international conference on computer vision}, 2021, pp. 7262--7272.

\bibitem{thisanke2023semantic_visionTransformer}
H.~Thisanke, C.~Deshan, K.~Chamith, S.~Seneviratne, R.~Vidanaarachchi, and D.~Herath, ``Semantic segmentation using vision transformers: A survey,'' \emph{Engineering Applications of Artificial Intelligence}, vol. 126, p. 106669, 2023.

\bibitem{chen2021transunet}
J.~Chen, Y.~Lu, Q.~Yu, X.~Luo, E.~Adeli, Y.~Wang, L.~Lu, A.~L. Yuille, and Y.~Zhou, ``Transunet: Transformers make strong encoders for medical image segmentation,'' \emph{arXiv preprint arXiv:2102.04306}, 2021.

\bibitem{cao2022swinunet}
H.~Cao, Y.~Wang, J.~Chen, D.~Jiang, X.~Zhang, Q.~Tian, and M.~Wang, ``Swin-unet: Unet-like pure transformer for medical image segmentation,'' in \emph{European conference on computer vision}.\hskip 1em plus 0.5em minus 0.4em\relax Springer, 2022, pp. 205--218.

\bibitem{lin2022ds_transunet}
A.~Lin, B.~Chen, J.~Xu, Z.~Zhang, G.~Lu, and D.~Zhang, ``Ds-transunet: Dual swin transformer u-net for medical image segmentation,'' \emph{IEEE Transactions on Instrumentation and Measurement}, vol.~71, pp. 1--15, 2022.

\bibitem{zhou2021nnformer}
H.-Y. Zhou, J.~Guo, Y.~Zhang, L.~Yu, L.~Wang, and Y.~Yu, ``nnformer: Interleaved transformer for volumetric segmentation,'' \emph{arXiv preprint arXiv:2109.03201}, 2021.

\bibitem{xie2017adversarial_segmentation}
C.~Xie, J.~Wang, Z.~Zhang, Y.~Zhou, L.~Xie, and A.~Yuille, ``Adversarial examples for semantic segmentation and object detection,'' in \emph{Proceedings of the IEEE international conference on computer vision}, 2017, pp. 1369--1378.

\bibitem{arnab2018robustness}
A.~Arnab, O.~Miksik, and P.~H. Torr, ``On the robustness of semantic segmentation models to adversarial attacks,'' in \emph{Proceedings of the IEEE conference on computer vision and pattern recognition}, 2018, pp. 888--897.

\bibitem{chen2020seqvat}
L.~Chen, W.~Ruan, X.~Liu, and J.~Lu, ``Seqvat: Virtual adversarial training for semi-supervised sequence labeling,'' in \emph{Proceedings of the 58th Annual Meeting of the Association for Computational Linguistics}, 2020, pp. 8801--8811.

\bibitem{kirillov2023SAM}
A.~Kirillov, E.~Mintun, N.~Ravi, H.~Mao, C.~Rolland, L.~Gustafson, T.~Xiao, S.~Whitehead, A.~C. Berg, W.-Y. Lo \emph{et~al.}, ``Segment anything,'' \emph{arXiv preprint arXiv:2304.02643}, 2023.

\bibitem{ma2023MedSAM}
J.~Ma and B.~Wang, ``Segment anything in medical images,'' \emph{arXiv preprint arXiv:2304.12306}, 2023.

\bibitem{gunasekar2023textbooks}
S.~Gunasekar, Y.~Zhang, J.~Aneja, C.~C.~T. Mendes, A.~Del~Giorno, S.~Gopi, M.~Javaheripi, P.~Kauffmann, G.~de~Rosa, O.~Saarikivi \emph{et~al.}, ``Textbooks are all you need,'' \emph{arXiv preprint arXiv:2306.11644}, 2023.

\bibitem{smith_nnforco_1999}
K.~A. Smith, ``Neural networks for combinatorial optimization: a review of more than a decade of research,'' \emph{INFORMS Journal on Computing}, vol.~11, no.~1, pp. 15--34, 1999.

\bibitem{bengio_mlforco_2021}
Y.~Bengio, A.~Lodi, and A.~Prouvost, ``Machine learning for combinatorial optimization: a methodological tour d’horizon,'' \emph{European Journal of Operational Research}, vol. 290, no.~2, pp. 405--421, 2021.

\bibitem{kotary_e2ecosntrainedoptlearning_2021}
J.~Kotary, F.~Fioretto, P.~Van~Hentenryck, and B.~Wilder, ``End-to-end constrained optimization learning: A survey,'' \emph{arXiv preprint arXiv:2103.16378}, 2021.

\bibitem{larson_urbanoperationresearch_1981}
R.~C. Larson and A.~R. Odoni, \emph{Urban operations research}.\hskip 1em plus 0.5em minus 0.4em\relax Englewood Cliffs, N.J., 1981, no. Monograph.

\bibitem{poganvcic_diffblackboxcombisolver_2019}
M.~V. Pogan{\v{c}}i{\'c}, A.~Paulus, V.~Musil, G.~Martius, and M.~Rolinek, ``Differentiation of blackbox combinatorial solvers,'' in \emph{International Conference on Learning Representations}, 2019.

\bibitem{elmachtoub_smartpredictoptimize_2021}
A.~N. Elmachtoub and P.~Grigas, ``Smart “predict, then optimize”,'' \emph{Management Science}, 2021.

\bibitem{mensch_ddp_2018}
A.~Mensch and M.~Blondel, ``Differentiable dynamic programming for structured prediction and attention,'' in \emph{International Conference on Machine Learning}.\hskip 1em plus 0.5em minus 0.4em\relax PMLR, 2018, pp. 3462--3471.

\bibitem{xie_DDP_2022}
H.~Xie, W.~Xu, and X.~Wu, ``A deep learning network with differentiable dynamic programming for retina oct surface segmentation,'' \emph{arXiv preprint arXiv:2210.06335}, 2022.

\bibitem{xie2023DDP}
H.~Xie, W.~Xu, Y.~X. Wang, and X.~Wu, ``Deep learning network with differentiable dynamic programming for retina oct surface segmentation,'' \emph{Biomedical Optics Express}, vol.~14, no.~7, pp. 3190--3202, 2023.

\bibitem{khalil_embeddinggraph_2017}
E.~Khalil, H.~Dai, Y.~Zhang, B.~Dilkina, and L.~Song, ``Learning combinatorial optimization algorithms over graphs,'' \emph{Advances in neural information processing systems}, vol.~30, 2017.

\bibitem{gasse_graphcnn_2019}
M.~Gasse, D.~Ch{\'e}telat, N.~Ferroni, L.~Charlin, and A.~Lodi, ``Exact combinatorial optimization with graph convolutional neural networks,'' \emph{Advances in Neural Information Processing Systems}, vol.~32, 2019.

\bibitem{wang_woundseg_2020}
C.~Wang, D.~Anisuzzaman, V.~Williamson, M.~K. Dhar, B.~Rostami, J.~Niezgoda, S.~Gopalakrishnan, and Z.~Yu, ``Fully automatic wound segmentation with deep convolutional neural networks,'' \emph{Scientific reports}, vol.~10, no.~1, pp. 1--9, 2020.

\bibitem{antonelli2022_msd}
M.~Antonelli, A.~Reinke, S.~Bakas, K.~Farahani, A.~Kopp-Schneider, B.~A. Landman, G.~Litjens, B.~Menze, O.~Ronneberger, R.~M. Summers \emph{et~al.}, ``The medical segmentation decathlon,'' \emph{Nature Communications}, vol.~13, no.~1, pp. 1--13, 2022.

\bibitem{brush1967history_isingmodel}
S.~G. Brush, ``History of the lenz-ising model,'' \emph{Reviews of modern physics}, vol.~39, no.~4, p. 883, 1967.

\bibitem{komodakis2010mrf}
N.~Komodakis, N.~Paragios, and G.~Tziritas, ``{MRF} energy minimization and beyond via dual decomposition,'' \emph{IEEE transactions on pattern analysis and machine intelligence}, vol.~33, no.~3, pp. 531--552, 2010.

\bibitem{hekaiming_residualconnection_2016}
K.~He, X.~Zhang, S.~Ren, and J.~Sun, ``Deep residual learning for image recognition,'' in \emph{Proceedings of the IEEE conference on computer vision and pattern recognition}, 2016, pp. 770--778.

\bibitem{tang2018normalized}
M.~Tang, A.~Djelouah, F.~Perazzi, Y.~Boykov, and C.~Schroers, ``Normalized cut loss for weakly-supervised cnn segmentation,'' in \emph{Proceedings of the IEEE conference on computer vision and pattern recognition}, 2018, pp. 1818--1827.

\bibitem{tang2018regularized}
M.~Tang, F.~Perazzi, A.~Djelouah, I.~Ben~Ayed, C.~Schroers, and Y.~Boykov, ``On regularized losses for weakly-supervised cnn segmentation,'' in \emph{Proceedings of the European Conference on Computer Vision (ECCV)}, 2018, pp. 507--522.

\bibitem{gibert2012graph_embedding}
J.~Gibert, E.~Valveny, and H.~Bunke, ``Graph embedding in vector spaces by node attribute statistics,'' \emph{Pattern Recognition}, vol.~45, no.~9, pp. 3072--3083, 2012.

\bibitem{hochreiter_gradientvanishing_1998}
S.~Hochreiter, ``The vanishing gradient problem during learning recurrent neural nets and problem solutions,'' \emph{International Journal of Uncertainty, Fuzziness and Knowledge-Based Systems}, vol.~6, no.~02, pp. 107--116, 1998.

\bibitem{residual2016}
K.~He, X.~Zhang, S.~Ren, and J.~Sun, ``Identity mappings in deep residual networks,'' in \emph{European conference on computer vision}.\hskip 1em plus 0.5em minus 0.4em\relax Springer, 2016, pp. 630--645.

\bibitem{ebrahimi_studyonresnet_2021}
M.~S. Ebrahimi and H.~K. Abadi, ``Study of residual networks for image recognition,'' in \emph{Intelligent Computing}.\hskip 1em plus 0.5em minus 0.4em\relax Springer, 2021, pp. 754--763.

\bibitem{sudre_gdl_2017}
C.~H. Sudre, W.~Li, T.~Vercauteren, S.~Ourselin, and M.~Jorge~Cardoso, ``Generalised dice overlap as a deep learning loss function for highly unbalanced segmentations,'' in \emph{Deep learning in medical image analysis and multimodal learning for clinical decision support}.\hskip 1em plus 0.5em minus 0.4em\relax Springer, 2017, pp. 240--248.

\bibitem{rick_covlossweight_2021}
R.~Groenendijk, S.~Karaoglu, T.~Gevers, and T.~Mensink, ``Multi-loss weighting with coefficient of variations,'' in \emph{Proceedings of the IEEE/CVF Winter Conference on Applications of Computer Vision}, 2021, pp. 1469--1478.

\bibitem{goffman1975_hyperplane_lebesgueMeasure}
C.~Goffman and G.~Pedrick, ``A proof of the homeomorphism of lebesgue-stieltjes measure with lebesgue measure,'' \emph{Proceedings of the American Mathematical Society}, vol.~52, no.~1, pp. 196--198, 1975.

\bibitem{borovkov2013_lebesgue_zero}
A.~A. Borovkov, ``Random variables and distribution functions,'' in \emph{Probability Theory}.\hskip 1em plus 0.5em minus 0.4em\relax Springer, 2013, pp. 31--63.

\bibitem{rao2016_lebesgue_zero}
C.~R. Rao and M.~M. Lovric, ``Testing point null hypothesis of a normal mean and the truth: 21st century perspective,'' \emph{Journal of Modern Applied Statistical Methods}, vol.~15, no.~2, p.~3, 2016.

\bibitem{pytorch_2019}
A.~Paszke, S.~Gross, F.~Massa, A.~Lerer, J.~Bradbury, G.~Chanan, T.~Killeen, Z.~Lin, N.~Gimelshein, L.~Antiga, A.~Desmaison, A.~Kopf, E.~Yang, Z.~DeVito, M.~Raison, A.~Tejani, S.~Chilamkurthy, B.~Steiner, L.~Fang, J.~Bai, and S.~Chintala, ``Pytorch: An imperative style, high-performance deep learning library,'' \emph{Advances in neural information processing systems}, vol.~32, pp. 8026--8037, 2019.

\bibitem{powers2020evaluation}
D.~M. Powers, ``Evaluation: from precision, recall and f-measure to roc, informedness, markedness and correlation,'' \emph{arXiv preprint arXiv:2010.16061}, 2020.

\bibitem{goodfellow_adversarialattack_2014}
I.~J. Goodfellow, J.~Shlens, and C.~Szegedy, ``Explaining and harnessing adversarial examples,'' \emph{arXiv preprint arXiv:1412.6572}, 2014.

\bibitem{isensee2021_nnunet}
F.~Isensee, P.~F. Jaeger, S.~A. Kohl, J.~Petersen, and K.~H. Maier-Hein, ``{nnU-Net}: a self-configuring method for deep learning-based biomedical image segmentation,'' \emph{Nature methods}, vol.~18, no.~2, pp. 203--211, 2021.

\end{thebibliography}

\begin{IEEEbiographynophoto}{Hui Xie}
	A senior AI data scientist in Optum Insight, USA. He received his PhD degree from the University of Iowa in 2022. His research interests include deep learning image processing and large language models for NLP.
	He developed the first in-house Generative AI platform in Optum and developed an LLM-based question/answer platform to support image and table output beside text.
\end{IEEEbiographynophoto}

\begin{IEEEbiographynophoto}{Weiyu Xu}
	Associate Professor in the Departments of Electrical and Computer Engineering at the University of Iowa. He received his Ph.D. degree from the California Institute of Technology, Pasadena, CA, in 2009. His research interests are signal processing, optimization, and high-dimensional data analytics. He is particularly interested in the role of optimization and high dimensional convex geometry in providing novel computational solutions to signal processing and high dimensional data analytics. Specific topics of my research include compressed sensing, low-rank matrix recovery, super-resolution, signal processing for communications, and optimizations for novel medical imaging and radiotherapy cancer treatment planning technologies.
\end{IEEEbiographynophoto}

\begin{IEEEbiographynophoto}{Ya Xing Wang, MD}
	Professor of Ophthalmology and a Glaucoma Specialist at Beijing Tongren Hospital. She received her MD degree from Capital Medical University and Heidelberg University. She obtained clinical training at Beijing Tongren Hospital and clinical research training at Moorfields Eye Hospital and UCL Institute of Ophthalmology. She also worked as a post-doc fellow at Devers Eye Institute in the United States. Her research interests include glaucoma, ophthalmic epidemiology, and optic disc biomechanics. She has co-authored 200+ papers in peer-reviewed journals and has been involved in several international collaborative studies. She now serves as an Editorial Board Member of EYE and the Asia Pacific Journal of Ophthalmology. She has been a Fellow of the Asia Pacific Professors of Ophthalmology (AAPPO) since 2021.
\end{IEEEbiographynophoto}

\begin{IEEEbiographynophoto}{John Buatti, MD}
	Professor in the Departments of Radiation Oncology, Neurosurgery, and Otolaryngology at the University of Iowa. He is also the Chair of the Department of Radiation Oncology, a position that he has held with distinction since 2001 when he established an independent department at the University of Iowa Hospitals and Clinics. He received his MD in Medicine, from Georgetown University, Washington, D.C, in 1986. Professor Buatti's research focuses on the use of imaging to improve outcomes for patients undergoing cancer therapy, radiation therapy in particular. In 2005, he opened Iowa's first Center of Excellence in Image Guided Radiation Therapy, which was innovative for its use of advanced 3D imaging including a 3T MR scanner and 40-slice respiratory-gated PET/CT scanner to better locate and treat tumors. As a clinician, he is a leading authority on the treatment of cranial malignancies. He is a prolific scholar having authored or co-authored more than 250 peer-reviewed publications. He has published 22 book chapters. With a lifelong commitment to improving the quality of cancer imaging in therapy, he served as the first chair of the Steering Committee for the Quantitative Imaging Network (QIN) and Chair of the Clinical Trials Design and Development Working Group, and was a member of the Coordinating Committee for QIN. He has served on the ASTRO Education Committee, as Co-Chair of the ASTRO IHE-RO (Integrating Health Enterprise-Radiation Oncology), and served on the ASTRO Science Council Steering Committee. He chairs the ASTRO task force on Theranostics.
\end{IEEEbiographynophoto}

\begin{IEEEbiographynophoto}{Xiaodong Wu}
	Professor in the Departments of Electrical and Computer Engineering and Radiation Oncology at the University of Iowa. He received his Ph.D. degree from the University of Notre Dame, in 2002. He is a researcher at the Iowa Technology Institute and the Iowa Institute for Biomedical Imaging. His research interests are primarily in biomedical computing and computer algorithms, with an emphasis on the development and implementation of efficient algorithms for solving problems that arise in computer-assisted medical diagnosis and treatment, biomedical image analysis, and bioinformatics. Professor Wu's research has been supported by the National Science Foundation (NSF), the National Institutes of Health (NIH), the American Cancer Society, and the University of Iowa. He is a recipient of the NSF CAREER Award and the NIH K25 Career Development Award.
\end{IEEEbiographynophoto}

\end{document}